\documentclass[preprint,12pt, authoryear]{elsarticle}
\usepackage[utf8]{inputenc}
\usepackage{geometry}
\usepackage{algorithm2e}
\RestyleAlgo{ruled} 
\usepackage{url}
\RequirePackage[pdftex,
      colorlinks={true},
      pdfstartview={FitH},
      bookmarks={true},
            pdfauthor={Author},
            pdftitle={Title},
            pdfsubject={Subject}
            plainpages=false,
            pdfpagelabels
            ]{hyperref}
\usepackage{graphicx}
\usepackage{xcolor}
\usepackage{booktabs}
\usepackage{subcaption}
\usepackage{adjustbox}
\usepackage{float}
\usepackage{multirow}
\usepackage{amssymb}
\usepackage[font=small,skip=0pt]{caption}
\usepackage{multirow}
\usepackage{comment}
\usepackage{threeparttable}
\usepackage[symbol]{footmisc}
\usepackage[acronym]{glossaries}
\makeglossaries

\newglossaryentry{ML}{
  type=\acronymtype, 
  name={ML}, 
  first={Machine Learning (ML)}, %\glsadd{BDGinfo}},
  description={Machine Learning}
}

\newglossaryentry{DT}{
  type=\acronymtype, 
  name={DT}, 
  first={Decision Tree (DT)},
  description={Decision Tree}
}

\newglossaryentry{RF}{
  type=\acronymtype, 
  name={RF}, 
  first={Random Forest (RF)},
  description={Random Forest}
}

\newglossaryentry{XGB}{
  type=\acronymtype, 
  name={XGB}, 
  first={Extreme Gradient Boosting (XGB)},
  description={Extreme Gradient Boosting}
}

\newglossaryentry{NB}{
  type=\acronymtype, 
  name={NB}, 
  first={Naive Bayes (NB)},
  description={Naive Bayes}
}

\newglossaryentry{SVM}{
  type=\acronymtype, 
  name={SVM}, 
  first={Support Vector Machine (SVM)},
  description={Support Vector Machine}
}

\newglossaryentry{NN}{
  type=\acronymtype, 
  name={NN}, 
  first={Neural Networks (NN)},
  description={Neural Networks}
}

\newglossaryentry{ZINB}{
  type=\acronymtype, 
  name={ZINB}, 
  first={Zero-Inflated Negative Binomial Regression (ZINB)},
  description={Zero-Inflated Negative Binomial Regression}
}

\newglossaryentry{LogR}{
  type=\acronymtype, 
  name={LogR}, 
  first={Logistic Regression (LogR)},
  description={Logistic Regression}
}
\newglossaryentry{MNL}{
  type=\acronymtype, 
  name={MNL}, 
  first={Multinomial Logit (MNL)},
  description={Multinomial Logit}
}

\newglossaryentry{LinR}{
  type=\acronymtype, 
  name={LinR}, 
  first={Linear Regression (LinR)},
  description={Linear Regression}
}

\newglossaryentry{NL}{
  type=\acronymtype, 
  name={NL}, 
  first={Nested Logit (NL)},
  description={Nested Logit}
}

\usepackage{lineno} %linenumbers
\usepackage{amsmath}

\begin{document}
\let\WriteBookmarks\relax
\def\floatpagepagefraction{1}
\def\textpagefraction{.001}

% \begin{highlights}
% \item Research highlights item 1
% \item Research highlights item 2
% \item Research highlights item 3
% \end{highlights}

\title{Understanding Transit Ridership in \textcolor{black}{an Equity Context}
%of Low-Income Carless Individuals 
through a Comparison of Statistical and Machine Learning Algorithms}

% \author{Author names are removed for the double anonymized review.}

\author[1]{Elnaz Yousefzadeh Barri$^*$}
\ead{elnaz.yousefzadeh@mail.utoronto.ca}
\cortext[*]{This paper is published in the journal of Transport Geography. Please refer to this link for the published manuscript: \href{https://doi.org/10.1016/j.jtrangeo.2022.103482}{https://doi.org/10.1016/j.jtrangeo.2022.103482} *Corresponding author}

\author[2]{Steven Farber}
\ead{steven.farber@utoronto.ca}

\author[3]{Hadi Jahanshahi}
\ead{hadi.jahanshahi@ryerson.ca}

\author[4]{Eda Beyazit}
\ead{beyazite@itu.edu.tr}

\address[1]{Department of Geography \& Planning, University of Toronto, Canada}
\address[2]{Department of Human Geography, University of Toronto Scarborough, Canada}
\address[3]{Data Science Lab, Ryerson University, Canada}
\address[4]{Department of Urban and Regional Planning, Istanbul Technical University, Turkey}

\journal{Journal of Transport Geography}

%%%%%%%%%%%%%%%%%%%%%%%%%%%%%%%%%%%
%%%%%%%%%%% ABSTRACT %%%%%%%%%%%%%%
%%%%%%%%%%%%%%%%%%%%%%%%%%%%%%%%%%%
\begin{abstract}
Building an accurate model of travel behaviour based on individuals' characteristics and built environment attributes is of importance for policy-making and transportation planning. Recent experiments with big data and Machine Learning (ML) algorithms toward a better travel behaviour analysis have mainly overlooked socially disadvantaged groups. Accordingly, in this study, we explore the travel behaviour responses of low-income individuals to transit investments in Greater Toronto and Hamilton Area, Canada, using statistical and ML models. We first investigate how the model choice affects the prediction of transit use by the low-income group. This step includes comparing the predictive performance of traditional and ML algorithms and then evaluating a transit investment policy by contrasting the predicted activities and the spatial distribution of transit trips generated by vulnerable households after improving accessibility. \textcolor{black}{We also empirically investigate the proposed transit investment by each algorithm and compare it with the city of Brampton's future transportation plan.} While, unsurprisingly, the ML algorithms outperform classical models, there are still doubts about using them due to interpretability concerns. Hence, we adopt recent local and global model-agnostic interpretation tools to interpret how the model arrives at its predictions. Our findings reveal the great potential of ML algorithms for enhanced travel behaviour predictions for low-income strata without \textcolor{black}{considerably} sacrificing interpretability.
\end{abstract}


\maketitle

%%%%%%%%%%%%%%%%%%%%%%%%%%%%%%%%%%%
%%%%%%%%%% INTRODUCTION %%%%%%%%%%%
%%%%%%%%%%%%%%%%%%%%%%%%%%%%%%%%%%%
% \linenumbers
\section{Introduction}
Accurate modelling of travel behaviour is an important component in transportation planning and travel demand management. For decades, extensive efforts have been devoted to identifying and improving methods in travel behaviour research, including such watershed moments as the derivation of discrete choice models, the shift from trip-based to activity-based models, and, more recently, experimentation with ``big data'' and Machine Learning (ML) methodologies. These technical and theoretical advancements go hand in hand with developing more nuanced understandings of the travel needs and revealing travel outcomes of different members of the population. From a social justice perspective, improving model accuracy is therefore vitally important to understand how different people respond to different types of changes in their transportation and land use environment and, accordingly, how to better plan for the needs of historically marginalized communities. Within justice-based transportation planning, travel behaviour models help researchers predict activity and travel behaviour outcomes associated with transit investments, which help planners evaluate the equity implications of different planning scenarios. Moving beyond the typical buffering exercises involved in US-based Title VI and Environmental Justice analyses, travel behaviour-based assessments examine how transit investments unlock potential for higher life quality. It can be done by forecasting behavioural responses such as changes in auto-ownership, transit mode share, and out-of-home activity participation rates.

Since the 1980s, most mode-choice problems have been addressed and modeled by traditional discrete choice models – e.g., multinomial logit. Their elegant mathematical formulation and convenient interpretation have made them popular~\citep{Lee2018}. However, ML algorithms demonstrated high accuracy in various research contexts and have received particular attention within the travel behaviour modelling literature~\citep{Xie2003, Hagenauer2017, Cheng2019}. An ML algorithm utilizes a set of rules or models to learn complex patterns within a dataset. It is very different from traditional discrete-choice models that are firmly grounded in microeconomic theory. However, ML's flexibility in dealing with non-linearity and capturing complex and previously undiscovered relationships between input and output variables makes them promising for modelling heterogeneous travel behaviour patterns. Notwithstanding their predictive performance, they are also referred to as ``black-box'' estimators since they lack behavioural-theoretic formalities that allow analysts to easily describe relationships between variables in the model. Nonetheless, travel behaviour researchers have been gradually adopting ML models for travel demand modelling, particularly mode-choice decisions \citep{Xie2003, Zhang2008, Cheng2019}. Within this burgeoning literature, to the best of our knowledge, there have been few efforts that examine the pros and cons of using ML models within transit equity research and planning. 

\textcolor{black}{Given the significance of using an accurate model to travel demand management, the main objective of this study is to examine how the model selection (e.g., statistical and ML algorithms) influences travel behaviour prediction, transportation planning, and policy-making for low-income households. We explore two main goals. First, we aim to investigate the potential of using ML models in predicting travel behaviour responses to transit investments among marginalized populations. As a result, we compare the predictive performance of the most commonly used ML algorithms with statistical models in transit use analysis. We apply statistical comparison to determine whether there is a significant difference in their prediction performances. Second, we aim to explore whether model selection affects the justice-based interpretation of scenarios. In other words, we study features affecting prediction outcomes and the possible interpretability of the ML models compared to the traditional ones. To this end, we utilize a sensitivity analysis to explore whether model selection would affect transit investment policies. We further compare the spatial distribution of the forecasted transit trips after transit improvements in the region for low-income carless families to assess the justice-based interpretation of the scenario. Moreover, we aim to discover how the potential forecasting accuracy benefits of ML approaches stack up against their potential drawbacks, i.e., they are not derived from behavioural theory and do not provide easily interpretable relationships between input and output variables.} 

%This study aims to assess the potential for using machine learning-based travel behaviour models to accurately predict travel behaviour responses to transit investments among marginalized populations. Moreover, we investigate the interpretability of Machine Learning (ML) models compared to the traditional approaches. 

%%%%%%%%%%%%%%%%%%%%%%%%%%%%%%%%%%%
%%%%%%%%%%%% LIT REV %%%%%%%%%%%%%%
%%%%%%%%%%%%%%%%%%%%%%%%%%%%%%%%%%%
\section{Literature review}\label{sec:litrev}

Travel behaviour analysis often aims to forecast travel demand in a city based on future transportation investments. When individual travel information is available, a disaggregated model easily captures travellers' behavioural patterns and predicts their decisions following changes in the transportation system. For transit equity analysis, the impacts of changes in transportation plans on different communities and equity outcomes of those policies can be evaluated by predicting travellers' participation in activities~\citep{Hodgson2003, Martens2016, Fransen2018, Allen2020, Yousefzadeh2021}. Accordingly, planners understand to what extent the outcome of policy options might be inequitable across different segments of the population. For instance, evaluating transit mode share or accessibility among people who have been historically marginalized enables transport planners and decision-makers to explore whether inequities are being redressed by proposed transit projects. Reviewing studies reveals that spatial attributes of the built environment \citep{cervero1997, Ewing2010}, socioeconomic characteristics of travellers \citep{turner1997, Dieleman2002}, and the level of transit service \citep{Taylor2003, Moniruzzaman2012} affect an individual's travel mode choice and behavioural pattern. Thus, it is desirable to have a disaggregated, comprehensive and accurate travel behaviour model that deals with complex behavioural attributes and predicts travellers' responses to the variations in the transportation system. Traditional discrete-choice or statistical models are mainly adopted to explore the travel mode choice of marginalized groups (e.g., ~\citep{Mercado2012, Yousefzadeh2021, Jiao2021}). However, to the best of our knowledge, the feasibility of using ML algorithms in predicting the travel mode choice of low-income individuals is yet to be investigated. Accordingly, here, we review transportation studies using either traditional or ML models to explore the travel behaviour patterns of any group of people. \textcolor{black}{In the next subsection, we summarize how conventional and ML models have been applied to trip count predictions and then explore studies comparing statistical and ML models in travel behaviour. Afterward, we explore ML regression models specifically utilized in the transit equity domain, followed by the trade-off between predictive performance and interpretability of ML models. Finally, we specify our paper's contribution to the literature.}

\subsection{\textcolor{black}{Statistical and ML models for count predictions}}
In evaluating transportation policies, both classification and regression algorithms are used to predict future travel demand and behavioural outcomes. Classification models, specifically discrete choice models, are extensively applied in mode-choice analysis for predicting a commuter's travel mode choice~\citep{Mercado2012, Jun2013, Abasahl2018, Jiao2021, Costa2022}. Notably, all these studies focus on conventional statistical models such as Binary and Multinomial Logistic Regression, Multinomial Logit model, Nested Logit model, and Mixed logit model. %Further, travel behaviour researchers use statistical count models for predictive modelling. 
On the other hand, a wide range of regression models, including Ordinary Least Square regression, Zero-Inflated count models, Hurdle, Negative Binomial, and Poisson, are popular in travel behaviour studies. These methods are utilized for estimating commuters' transit trip frequency \citep{Yousefzadeh2021, bocker2017, Legrain2016}, \textcolor{black}{variations in bicycle ridership counts~\citep{Roy2019},} the relationship between residential self-selection and non-work trips \citep{Chatman2009}, built environment and trip generation association \citep{Zhang2019}, bus fare evasion rates \citep{Guarda2016}, and peak-car phenomenon (leveling off car travel) \citep {Kamruzzaman2020} as they are easy to apply and interpret. 
%Accordingly, investigating individuals' travel patterns baesd on their trip numbers is one of the main applications of count models. 
The implementation and interpretation of statistical models are straightforward, although they fail to handle the nonlinear relationship between variables. 

Typically, the advantage of ML models is to learn and model the intricate interactions between a dependent and a set of independent features~\citep{Xie2003, Cheng2019}, making them suitable for modelling travel behaviour. 
\textcolor{black}{Accordingly, many studies have recently used ML algorithms including Decision Tree (DT), Random Forest (RF)~\citep{Cheng2019, Yan2020}, Extreme Gradient Boosting (XGB)~\citep{Wang2018,Shao2020}, Neural Networks (NN)~\citep{Xie2003}, and Support Vector Machine (SVM)~\citep{Zhang2008, Zhou2019} for travel demand and mode choice predictions. \citet{Shao2020} investigate the nonlinear relationship between land use and metro ridership at the station level using the XGB model. %They examine the effects of land use and several station-area demographic variables of each station on daily station ridership. 
In modelling travel demand for ride-sourcing, \citet{Yan2020} estimate the number of ride-sourcing trips using the RF algorithm and the traditional multiplicative model and then compare their predictive performance in forecasting the future number of ride-sourcing trips.}

\textcolor{black}{Several studies have compared the predictive performance of ML models with that of statistical models, and most of them have indicated that ML models are promising tools compared to statistical models (e.g., \citep{Xie2003, Hagenauer2017, Cheng2019, Zhou2019, Yan2020, Chen2021}). Focusing on travel mode choice modelling,~\citet{Hagenauer2017},~\citet{Cheng2019},~\citet{Zhou2019}, and \citet{Zhao2020} use multiple ML and statistical classifiers to compare their predictive capability and behavioural analysis. Most of these studies focus on the classification problem rather than regression. However, there are a few studies constructing a regression problem using both statistical and ML models. For instance, \citet{Kim2020} have adopted mixed models to achieve more accurate travel demand prediction models. They consider the travel demand for on-demand ride-hailing services and develop a new framework integrating the Linear Regression model (LinR) and Long-term Short Memory model to predict the potential travel demand. In a similar approach, \citet{Chen2021} propose a hybrid model incorporating Geographically Weighted Regression (GWR) in the structure of RF to consider spatial heterogeneity and explore the complex association between built environment variables and bus-metro transfers (trip generations from bus to metro). Then, they compare the predictive performance of Multiple Linear Regression, SVM, RF, GWR, and their hybrid model to examine their suggested model’s advantages over other models. Therefore, ML models allow us to capture complex and nonlinear variable relationships and overcome the limitations of statistical modelling methodologies. Our approach differs from the aforementioned study in that we extensively compare ML and statistical models in the context of regression problems with an equity consideration. Moreover, we leverage statistical tests and interpretability tools to shed light on the differences among models and their possible explanations.}

\subsection{\textcolor{black}{ML applications in transit equity studies}}
\textcolor{black}{Over the decades, investigating transit services with an equity lens has become a crucial concern. Recently, some researchers have begun to adopt different ML techniques to address regression, classification, or clustering problems for their transportation equity studies. For instance, \citet{Tran2022} explore the travel experience and behavioural responses of transit riders before and during the COVID-19 pandemic. They utilize Twitter data for sentiment analysis as an ML algorithm to understand how a significant transit disruption may aggravate the vulnerabilities of transit-dependent users. \citet{Jiao2022} use the RF model to analyze e-scooter ridership in poorly transit-served neighbourhoods and address the inequality in transportation supply. %They predict the number of e-scooter riders and determine which socioeconomic and health factors are important in the final prediction. 
In measuring the vulnerability of households to transport energy burdens, \citet{Liu2022} propose a new framework to quantify transport-related energy poverty. They apply linear regression and several non-linear methods (e.g., DT, XGB, RF, and NN) to estimate households' average fuel consumption. They find that low-income households are more at risk of higher transport fuel costs compared to their counterparts. Among transportation studies, substantial studies in travel demand prediction (e.g.,\citep{Yan2020, Kim2020}), mode choice modelling (e.g.,\citep{Hagenauer2017, Zhao2020}), and traffic predictions (e.g.,\citep{Cai2016,Liu2019, Cui2020}) have adopted different ML algorithms. However, the potential of ML regressors in formulating the travel behaviour and transit use of low-income people and comparing their predictive performance with that of statistical models are not yet fully explored. Therefore, there is still a need to better investigate ML applications in the transit equity context in the literature. Moreover, interpretation of models is likely more important within an equity context compared to more basic travel demand forecasting, as researchers and planners are interested in the specific policy levers they can use to improve the well-being of marginalized populations.}

\subsection{Interpretable ML models}
With advances in ML techniques, their applications were applied in the transportation field. Despite their high predictive power, there is still a lack of research using ML techniques due to interpretability concerns~\citep{rudin2019, Koushik2020}. Even more important than the model performance is the interpretability of algorithms. 
\textcolor{black}{After knowing the probability of an event's occurrence, we need to discover how the prediction is made. Although an accurate predictive model can enhance the final decision, understanding the rationale behind the suggestions leads to a better insight into the problem.} 
Researchers generally use inherently interpretable models rather than ML models that are assumed to be black boxes. We note that the output of most ML models is not directly interpretable \citep {Zhao2020}. However, to decipher the way an ML model arrives at its conclusion, several interpretation techniques have been recently introduced. Unlike intrinsic interpretability, model-agnostic interpretability tools are employed following estimation of an ML model in a \textit{post hoc} analysis. The flexibility of these post hoc interpretation methods lets researchers use any ML model in different fields. 
\textcolor{black}{Applying interpretation tools to an algorithm is a way to understand the rationale behind the decision and justify the predicted outcome. Post hoc interpretability approaches uncover the effects of independent features on the response variable and interpret their influence to propose appropriate policy implications. Therefore, they summarize the behaviour of a model and explain how important a predictor is for the final decision enabling planners to identify key variables for effective decision making. More specifically, such explanations can provide enough evidence to implement policies for a given scenario.} 

When it comes to the granularity level of interpretability, global and local interpretation techniques describe the aggregated behaviour of the model and each individual or group, respectively \citep {Du2019, Molnar2020}. Feature importance \citep{Breiman2001, Hagenauer2017}, Partial Dependence Plot (PDP) \citep{Friedman2001}, and Accumulated Local Effects (ALE) \citep{Molnar2020} are examples of global interpretability tools. They explain the effect of each feature on the average prediction of a model. Conversely, Individual Conditional Expectation (ICE) \citep{Goldstein2015}, Shapley value (SHAP) \citep{Lundberg2017}, and Local Interpretable Model-agnostic Explanations (LIME) \citep{Ribeiro2016} explain how a model predicts an output for an individual \citep{Molnar2020}. Hence, contrary to the popular belief of black-box ML models, there are a plethora of post hoc tools that have the potential to address this limitation. In this study, we explore the ML models' interpretability in predicting the responses of marginalized individuals to transit improvements. \textcolor{black}{To the best of our knowledge, most studies in the mode choice analysis literature concentrate on global interpretation, whereas we report both global and local interpretability of ML models, especially in the context of transit equity.}

\subsection{Contribution of this study}
\textcolor{black}{In this study, we compare the performance of several ML algorithms with statistical models in transit equity analysis.
Our models predict the individuals' transit-use rates with regression approaches. We define three tasks to evaluate the differences between the models' performances. First, we compare the prediction performance of the number of transit trips made by low-income people using five ML and three conventional algorithms. To evaluate whether the difference in the predictive performance of the models is statistically significant, we use Friedman Aligned Ranks test. Then, Bergmann-Hommel post hoc analysis is applied to explore differences between each model pair and determine which algorithm differs from the others. Second, we evaluate a scenario with enhanced transit accessibility throughout the region for low-income carless families to investigate how people living in vulnerable households respond to transit investment policy. We compare the spatial distribution of the new transit trips predicted by all models after transit improvements in the region. Third, we leveraged model-agnostic interpretability tools to examine whether we can provide insight into ML models, often considered to be black boxes in the literature.}  

\textcolor{black}{Comparing the predictive performance of models through learning from data is of importance. However, interpretability adds another layer helping researchers realize how a model works and arrives at its conclusion. Given the growing importance of ML interpretability, several methods are introduced for this goal. Accordingly, in this study, we discuss the feasibility of the interpretability of ML models since this post hoc procedure is valuable for policy recommendations and future travel demand management. We consider model-agnostic interpretation tools, which are more flexible and can be generalized to any model type. Finally, we evaluate their interpretability using both global and local interpretability techniques.}

Table~\ref{table1:literature review} lists travel behaviour studies that use ML algorithms and conduct a comparative analysis with statistical models. Most of the studies focus on mode choice modelling as a classification task. Moreover, only one of them uses statistical tests to compare the significance of the differences among models \citep{Hagenauer2017}. Not all travel behaviour studies have used a validation technique for the unbiased train-test split -- e.g., cross-validation (CV) -- to compare the algorithms' performances. More importantly, the local interpretability of ML models is mainly disregarded. These works mostly applied global interpretability tools and ignored the importance of local interpretation. The novelty and difference of our paper compared to the existing literature are summarized in Table~\ref{table1:literature review}. \textcolor{black}{Besides the technical difference of our work, we take into account the transit ridership of vulnerable groups, making it a different unexplored domain for ML vs. statistical model comparison.}

\begin{table}[!ht]
    \centering
    \caption{Contribution of our study compared to the comparative analyses in the mode choice analysis literature}
    \resizebox{\textwidth}{!}{
\begin{tabular}{l|lcccccc}
    \toprule
    \multirow{2}{*}{\textbf{}} &
    \multirow{2}{*}{\textbf{Unit of analysis}} &
    \multicolumn{2}{c}{\textbf{Supervised Learning}} &
    \multirow{2}{*}{\textbf{\begin{tabular}[c]{@{}c@{}}Statistical \\Test$^\S$\end{tabular}}} & 
    \multirow{2}{*}{\textbf{CV}} & 
    \multicolumn{2}{c}{\textbf{Interpretability}} %& %\multirow{2}{*}{\textbf{\begin{tabular}[c]{@{}c@{}}Running\\ Time \\Report \end{tabular}}} & \multirow{3}{*}{\textbf{\begin{tabular}[c]{@{}c@{}}Data\\ Size \\ (Individuals)\end{tabular}}} 
    \\ 
    && \textbf{classification} & \textbf{regression} & %\textbf{\begin{tabular}[c]{@{}c@{}}threshold\\ dep.\end{tabular}} & \textbf{\begin{tabular}[c]{@{}c@{}}threshold\\ indep.\end{tabular}} &  &
    && \textbf{global} & \textbf{local} %& & 
    \\
    \midrule
    \textbf{\citet{Xie2003}} & Mode choice analysis & \checkmark &  &  &  &\checkmark & %& \checkmark & 15,064 
    \\
    \textbf{\citet{Zhang2008}} & Mode choice analysis &\checkmark  & &  &  & & %& & NA (5,029 trips) 
    \\
    \textbf{\citet{Hagenauer2017}}\textsuperscript{*} & Mode choice analysis &\checkmark  &   & \checkmark & \checkmark & \checkmark &  %& & 69,918 
    \\
    \textbf{\citet{Wang2018}} & Mode choice analysis & \checkmark &  &  &\checkmark  &\checkmark  &  %& & NA (51,910) trips 
    \\
    \textbf{\citet{Cheng2019}}\textsuperscript{*} & Mode choice analysis &\checkmark  &   &  &  &\checkmark  & %& \checkmark  & 2,991
    \\
    \textbf{\citet{Zhou2019}}\textsuperscript{*} & Mode choice analysis & \checkmark &   &  & \checkmark &  &  %& & 30,000
    \\
    \textbf{\citet{Zhao2020}}\textsuperscript{*} & Mode choice analysis & \checkmark  &   &  & \checkmark & \checkmark  &  %& & 1,163
    \\
    \textbf{\citet{Yan2020}} & Ride-sourcing trips &   & \checkmark  &  & \checkmark & \checkmark  &  
    \\
    \textbf{\citet{Chen2021}} & Bus-metro transfers &   & \checkmark  &  & \checkmark & \checkmark  &  
    \\
    \midrule
    \textbf{Our study}\textsuperscript{*} & Transit trip generation &   & \checkmark  & \checkmark & \checkmark & \checkmark & \checkmark %& \checkmark & 149,177
    \\
    \bottomrule
    \multicolumn{8}{l}{\begin{tabular}[c]{@{}l@{}}
    \small 
    {\footnotesize $^\S$ \small Using a statistical test to compare the significance of the difference in models' performances.}\\
    {\footnotesize \textsuperscript{*} \small These studies have compared more than two different ML algorithms with statistical models.}
    \end{tabular}}
    \end{tabular}
}
    \label{table1:literature review}
\end{table}

%%%%%%%%%%%%%%%%%%%%%%%%%%%%%%%%%%%
%%%%%%%%%$%%% DATA  %%%%%%%%%%%%%%%
%%%%%%%%%%%%%%%%%%%%%%%%%%%%%%%%%%%
\section{Data}\label{sec:data_and_methods}
In this study, we discuss how different ML techniques can improve prediction performance in transportation analysis projects \textcolor{black}{among low-income communities. To this end, we evaluate the travel behaviour of low-income households with a total income of less than \$40k per year.}

\subsection{Data source}
The data source used for this evaluation comes from the 2016 Transportation Tomorrow Survey (TTS), a large sample households travel survey including a one-day household travel diary. The sampling rate of this survey is 5\% of households in the Toronto region, except a 3\% sampling rate for Hamilton. Therefore, a set of expansion factors are considered to provide a reliable population estimation~\citep{TTS}. \textcolor{black}{We note that there might be sampling bias in the survey itself, which is partially addressed by weighting adjustments. By correcting for representation by dwelling type, family size, age, and gender, the data expansion procedure may better reflect additional variables (car ownership or employment status). We utilize the data with the caveat that there might still be additional aspects that cannot be found or corrected using the expansion factor.}

\textcolor{black}{Our main dataset for the study includes households with a total income level of less than \$40k per year, known as low-income households in our study area.} Further, we limit our analysis to only individuals aged 18 years and older, i.e., assumed to be autonomous in their decisions regarding residential location, vehicle ownership, and daily travel mode choice. Furthermore, we consider the individual trips which start from and end in GTHA regions and remove people with no trips or the number of trips greater than 25.

Table \ref{table: description} provides a description summary of explanatory variables, including individual socioeconomic characteristics, their trip information, and local built environment attributes. Notably, all figures provided in the table are expanded. Our dataset consists of 22,213 individuals of \textcolor{black}{low-income households with} a total of 61,539 trips, which are expandable to 531,406 individuals and 1,466,099 trips. \textcolor{black}{These attributes are all checked for possible multicollinearity, and we do not observe any significant correlation among them. To make it comparable with the previous works using the same dataset~\citep{Yousefzadeh2021, Allen2020}, we do not apply any normalization or standardization to the data.} Regarding the numeric attributes, \textcolor{black}{a gravity-based} transit accessibility measure is calculated as the total number of reachable jobs within the estimated travel time using the impedance function. \textcolor{black}{We used the calculated transit accessibility measure in a recent study in GTHA~\citep{Allen2019}. It is computed as}

\begin{equation}
    {A}_{i} = \sum_{j=1}^J O_j f(t_{ij})
    \label{eq:transit_acc}
\end{equation}
\textcolor{black}{where $A_i$ is the transit accessibility score in location $i$, $O_j$ defines the number of reachable jobs in zone j, $f(t_{ij})$ is the impedance function, and $t_{ij}$ is the travel time between $i$ and $j$.} This impedance function defines a decrease in attracting further away from jobs by increasing the travel time. To calculate the transit travel time, walking time to and from transit stops, waiting time for the transit, in-vehicle travel time by transit, and transferring time using GTFS and OpenStreetMap data are considered~\citep{Allen2020}. \textcolor{black}{This impedance function is as follows}

\begin{align}
    {f}(t_{ij}) = \begin{cases}
    180 (90+ t_{ij})^{-1}-1 &  t_{ij} < 90 \\
    0                     &  \text{otherwise}
    \end{cases}
    \label{eq:impedance_function}  
\end{align}

\textcolor{black}{It ranges between 0 and 1. For the median duration of travel time, which is 30 min, it gives 0.5. In 90 min which is the maximum travel time, the weight is 0.} The population density in each Dissemination Area and business density is extracted from the 2016 Canadian Census and the Canadian business registry, respectively. The intersection density equals the number of 3-way and more intersections per square kilometer.
       
\begin{table}[!ht]
    \centering
    \caption{Description of explanatory variables for people with their income less than $\$40k$ ($n=22,213; \mathcal{N}=531,406$)}
    \resizebox{0.70\textwidth}{!}{
    \begin{tabular}{lrr}
        \toprule
        \textbf{Categorical Variables} & &    \textbf{Proportion} \\
         \hline
         \textbf{Individual attributes}  &  & \\
         \qquad  {Age group\textsuperscript{*}} & & \\
         \qquad  \quad 18-25 & & 16.4\% \\
         \qquad  \quad 26-35 & & 15.5\% \\
         \qquad  \quad 36-45 & & 15.7\% \\
         \qquad  \quad 46-55 & & 16.1\% \\
         \qquad  \quad 56-65 & & 14.8\% \\
         \qquad  \quad 65+ & & 21.4\% \\
         \qquad  \quad Missing & & 0.1\% \\
         \qquad {Gender} &  &  \\
         \qquad  \quad Female & & 54.6\% \\
         \qquad  \quad Male & & 45.4\% \\
         \qquad  {Vehicles per adult} &  &  \\
         \qquad  \quad VA = 0 & & 29.2\% \\
         \qquad  \quad 0 \textless VA \textless 0.5 & & 12.3\% \\
         \qquad  \quad VA = 0.5 & & 25.9\% \\
         \qquad  \quad 0.5 \textless VA \textless 1 & & 6.4\% \\
         \qquad  \quad VA = 1 + & & 26.3\% \\
         \qquad {Having transit pass} &  &  \\
         \qquad  \quad Yes & & 29.8\% \\
         \qquad  \quad No & & 70.2\% \\
         \qquad {Having driving license} &  &  \\
         \qquad  \quad Yes & & 75.0\% \\
         \qquad  \quad No & & 25.0\% \\
         \qquad {Free parking at workplace} &  &  \\
         \qquad  \quad Yes & & 29.0\% \\
         \qquad  \quad No & & 11.5\% \\
         \qquad  \quad NA & & 59.6\% \\
         \midrule
         \textbf{Numeric Variables} & \textbf{$\mu$} & \textbf{$\sigma$} \\
         \hline
         \textbf{Trip information} &  &  \\
         \qquad \begin{tabular}[c]{@{}l@{}}Measure of accessibility to jobs\\ using a gravity function (transit commute)\textsuperscript{§} \end{tabular} & 217,355.2 & 271,689.9 \\
         \qquad Distance of mandatory trips (km) & 18.4 & 34.2 \\
         \qquad Distance of discretionary trips (km) & 7.0 & 16.2 \\
        \textbf{Built environment attributes} &  &  \\
         \qquad Population density (per person)\textsuperscript{‡} & 9,341.6 & 14,501.2 \\
         \qquad Business density (per person)\textsuperscript{‡} & 700.9 & 1,604.1 \\
         \qquad Intersection density (per person)\textsuperscript{‡} & 57.8 & 75.0 \\
        \bottomrule 
        \multicolumn{3}{l}{\begin{tabular}[c]{@{}l@{}} {\footnotesize \textsuperscript{*} We used this variable as a numeric feature in the modeling; however, we report it in a} \\ {\footnotesize categorical form for demonstration purpose.}\end{tabular}} \\
        \multicolumn{3}{l}{\begin{tabular}[c]{@{}l@{}} {\footnotesize \textsuperscript{§} Gravity-based accessibility to jobs by transit estimates the total number of reachable} \\ {\footnotesize jobs found in census tract from each dissemination area.}\end{tabular}} \\
        \multicolumn{3}{l}{\begin{tabular}[c]{@{}l@{}} {\footnotesize \textsuperscript{‡} Local built environment attributes of travelers are from the weighted sum of values} \\ {\footnotesize normalized by area in each dissemination area.}\end{tabular}} \\
    \end{tabular}
    }
    \label{table: description}
\end{table}

%\subsection{Experiment design}
 \subsection{Data preparation} \label{CV}
An unbiased model evaluation process has two folds: training the model on a training set and evaluating it on an unseen dataset, called a test set. Accordingly, the dataset is split arbitrarily into two parts, namely the training and test sets. It mitigates the overfitting problem. Nevertheless, this random split can produce bias~\citep{ElNaqa2015}. This study uses the stratified $k$-fold cross-validation (CV) technique to alleviate the bias issue. It is the most common approach in which the entire available dataset is divided into exclusive $k$ subsets of almost equal size. The ``stratified'' CV is adopted in case imbalanced classes exist. Hence, in our folds, we maintain the same ratio between non-transit and transit riders; otherwise, the majority class, i.e., non-transit users, might be overrepresented in some folds. In this technique, the model iterates the training and validation sets $k$ times where a subset is selected as a test set and the remaining ones as the training set.

In this study, we estimate each algorithm's performance using 10-fold cross-validation. In each iteration, we fit a model on nine folds and test it on the remaining one. After ten iterations, we record ten independent performance scores for each model. We report the average performance estimation during the CV. We compare algorithms' performance by their mean and standard deviation of this process. Finally, we utilize non-parametric statistical tests to assess the differences in model performances. All the comparisons are reported at the significance level of $\alpha = 0.05$.

%%%%%%%%%%%%%%%%%%%%%%%%%%%%%%%%%%%
%%%%%%%%$%%% METHODS %%%%%%%%%%%%%%
%%%%%%%%%%%%%%%%%%%%%%%%%%%%%%%%%%%
\section{Methods and algorithms}
%\subsection{Methods and algorithms}
The overall aim of this study is to compare ML algorithms with statistical methods to investigate their predictive performance \textcolor{black}{and their potential influence on transit investment policies. In this analysis, the number of transit trips per person is our dependent variable, whereas independent variables consist of sociodemographic, trip attributes, and built environment factors.} 

In this study, we apply five of the most commonly used ML algorithms in travel behaviour studies, including Decision Trees (DT), Random Forest (RF), eXtreme Gradient Boosting (XGB), Support Vector Machine (SVM), and Neural Networks (NN); and statistical methods such as LinR, Zero-Inflated Negative Binomial Regression (ZINB), and Hurdle models as baselines.

\textcolor{black}{We adopt a three-step approach to achieve our research objective. First, we estimate the predictive performance of each algorithm using various evaluation measures. To statistically examine the significant difference in each performance among all algorithms, we use a Friedman Aligned ranks test. On top of that, Bergmann-Hommel post hoc analysis was employed to make a pairwise comparison between models~\citep{Derrac2011}. In the second step, we used a sensitivity analysis to explore how a model selection may influence different predictions, spatial distribution, and planning policies. To check the difference in the spatial pattern of the predicted new trips, we applied the SPAtial EFficiency metric for each map~\citep{Demirel2018}}. In this section, a brief discussion about the specifications of each algorithm and various metrics used to compare their predictive performance are provided. \textcolor{black}{Further, we explain the statistical tests and measures used for this comparison.} 

\textcolor{black}{In selecting a predictive algorithm, the interpretability of a model can be as important as the model's predictive performance. In the last step, we investigate whether there is indeed a trade-off between predictive performance and interpretability of models. We discuss the feasibility of the interpretability of ML models. We consider model-agnostic interpretation methods that are more flexible and can be generalized to any model type. Accordingly, global and local interpretability tools are applied for the best performing ML model according to its predictive performance to understand how a model predicts and which factors and to what degree contribute to its prediction. A discussion about global and local interpretability techniques are provided in this section.} All these modelling and analyses are done in \texttt{R} and \texttt{Python} programming environments.

\subsection{Statistical models}
Regression analysis, one of the widely used statistical modelling approaches, explores the dependency of a predictor and responses. It helps understand a causality relationship between two or more continuous variables. Linear Regression (LinR) is one of the simplest regression models with a linearity assumption. It is applied when the dependent variable is continuous. 

One of the other well-known count regression models is negative binomial regression, in which the dependent variable follows the negative binomial distribution. It can be utilized if the dataset has overdispersed count outcome –- that is, the variance of the data is equal to or greater than its mean. Zero-Inflated Negative Binomial Regression (ZINB) and Hurdle model are examples of zero-inflated count models, particularly used in dealing with excessive zeros in the data. In the zero-inflated model, excessive zeros are divided into ``structural'' and ``sampling'' zeros. Sampling zeros come from the unusual Poisson or negative binomial distribution, assumed to be generated by chance. On the other hand, structural zeros are observed by non-risk groups who structurally are a source of zero~\citep{Hu2011, hua2014}. Either a mixture or a two-part modelling type, both ZINB and Hurdle models consist of two processes and deal with two types of distributions: zeros and counts. In the first process, a binomial model is utilized to estimate the probability of zeros versus non-zeros~\citep{Zuur2009, cameron1998}. In this step, the zero-inflated model assumes zeros both as structural and sampling zeros, while the hurdle model assumes all zeros as structural ones then formulate a pure mixture of zero and positive (non-zero) models~\citep{Hu2011, hua2014}. Therefore, their difference lies in the way they treat different types of zeros.

Another distinction between ZINB and the Hurdle model is in the second step, the count model portion. A truncated-at-zero count model is implemented for the count portion of the Hurdle model, while a negative binomial is used for the count portion of the zero-inflated model \citep{Zuur2009}.   
%Naive Bayes (NB) one of the simplest learning algorithm uses Bayes' rules.
All of the statistical models are derived based on strict assumptions about the data. Violation of these assumptions leads to inefficient and/or biased estimations. Conversely, there are no hypotheses or restrictive considerations in ML methods.

\subsection{Machine Learning models}
A machine learning (ML) algorithm as a computational process adjusts and improves its architecture through learning from the environment. This learning process stemmed from using and experiencing input data to achieve a required output. Since this training process constructs a fundamental part of this technique, most ML methods are classified based on their learning into two broad categories, \textit{supervised} and \textit{unsupervised learning} \citep{ElNaqa2015}. In \textit{supervised learning} methods, the dataset labels are known, helping the learning process to predict the outcome of new, unseen data.

A Decision Tree (DT) consisting of branches, decision nodes, and terminal leaves comes from the recursively partitioned feature space of the training set. The purpose of these tree-like structures, such as CART \citep{Breiman1984}, is to construct disjoint subnodes through a set of decision rules according to features. In a fully developed tree, this splitting process of the dataset iterates until all possible decision boundaries are tested, finally terminating at a terminal leaf, i.e., a homogeneous subnode. The impurity level of each decision node and the expected entropy reduction is computed to quantify the best split \citep{Wang1984, ElNaqa2015}. The terminal node in DT regressors is
the numeric estimated value for the dependent feature. To avoid overfitting problems and improve the predictive performance of models, ensemble techniques such as bagging \citep{Breiman1996} and boosting have been proposed. Random Forest (RF) is an ensemble ML algorithm that aggregates a collection of DTs with a random selection of features independent of previous attributes in each split \citep{Breiman2001}. Similar to all bagging models, the ultimate prediction result of RF is taken based on a majority vote of successive trees. Another tree-based ensemble model is Extreme Gradient Boosting (XGB), a scalable gradient tree boosting system~\citep{Chen2016}. It constructs consecutive weak trees by incrementally adding a new DT to prevent overfitting issues and improve predictive performance for hard-to-predict instances~\citep{Friedman2001}. 
 
While tree-based algorithms are formed of branches and leaves, Neural Networks (NN), considered ``black box'' algorithms, consist of layers and neurons (nodes). NN models are triggered by feeding input data and going through the activation functions to estimate the output values (nodes) using the sum of weighted connections in hidden layers. The most widely used way of optimizing weights is the backpropagation method, in which the weights are iteratively updated to minimize the total loss. On the other hand, Support Vector Machine (SVM) is a supervised learning model that can be divided into linear and non-linear models. To classify a linear dataset, a hyperplane is determined to define a boundary with a maximum margin between two classes and separate the data points \citep{Suthaharan2016}.
This hyperplane in the transformed space is the line close to support vectors —i.e., data points on the margin— with a maximum margin between those vectors \citep{Cortes1995, Vapnik2013}.

\subsection{Performance metrics}\label{sec:performance-metrics}
One of the major steps in model selection is evaluating the algorithm's performance. In predictive models, the estimation of predictive performance reflects how well the algorithm performs on unseen data. Therefore, selecting the best-performing model requires an approach to compare and rank the model's performance. In this study, multiple performance metrics are adopted to evaluate each technique. 
The performance of regressors is evaluated through R-Squared, Root Mean Squared Error (RMSE), Median Absolute Error (MedAE), and Root Relative Squared Error (RRSE) metrics.

To evaluate the predictive quality of different regression models, we compare the goodness-of-fit of each regressor by measuring R-squared. Other performance metrics are calculated based on the loss functions of the predicted errors, such as RMSE, MedAE, and RRSE. Unlike R-squared, the lower values for them are desirable. \textcolor{black}{Table~\ref{tab:perf_metrics} shows the performance metrics used in this study, where $n$ is the total number of observations, $y_i$ is the values of the dependent variable for observation $i$, $\hat{y}_i$ is its predicted value, and $\bar{y}$ is the average values of the dependent variable.}

\begin{table}[!ht]
\renewcommand{\arraystretch}{1.8}
\setlength{\tabcolsep}{20pt}

\centering
    \caption{\textcolor{black}{Performance metrics used for the regression problem}}\label{tab:perf_metrics}
    \begin{tabular}{ll}
    \toprule
    \textbf{Metric} & \textbf{Formula} \\
    \midrule
    $\text{R}^2$ & $1-\frac{\sum_{i=1}^{n}(y_{i}-\hat{y}_{i})^{2}}
        {\sum_{i=1}^{n}(y_{i}-\bar{y})^{2}}$ \\
    RMSE & $\sqrt{\sum_{i=1}^{n}\frac{(y_{i}-\hat{y}_{i})^2}{n}}$ \\
    MedAE & $\text{median}_{\forall i} (|{y_{i}-\hat{y_i}}|)$ \\
    RRSE & $\sqrt{\frac{\sum_{i=1}^{n}(y_{i}-\hat{y}_{i})^2}{\sum_{i=1}^{n}(y_{i}-\bar{y})^2}}$ \\
    \bottomrule
    \end{tabular}
\end{table}

%%%%%%%%%%%%%%%%%%%%%%%%%%%%%%%%%%%%%%%%
%%%%%%%%%%%%%%%% tests %%%%%%%%%%%%%%%%%
%Friedman Aligned ranks test and Bergmann-Hommel post hoc analysis
\subsection{\textcolor{black}{Non-parametric statistical tests for comparing multiple groups}} 
\textcolor{black}{We employ the Friedman test as a non-parametric statistical analysis with the block design that uses the ranked values to perform a comparison between more than two models. This test is used to determine if there is a statistically significant difference between the prediction performance of at least two algorithms. In this study, we employ the Friedman Aligned Ranks test, including the advanced ranking approach. The method of aligned ranks is suggested when the size of our dataset is small (i.e., the data does not follow the normal distribution)~\citep{Garcia2010, Derrac2011}. In our case, as we employ 10-fold cross-validation and have only 10 values to be compared, only a non-parametric test can be used. On the other hand, as the folds (datasets) remain the same for all iterations on different algorithms, we have to use a block design (i.e. a paired test) to compare the performance of the algorithms. The Friedman test facilitates such a pairwise comparison of the algorithms given each fold.}

\textcolor{black}{In this calculation, the average performance obtained from all algorithms in each fold for each metric is computed. Then, the difference between the performance score of each algorithm in each fold and the mean value of the same fold is calculated. Finally, these aligned scores, obtained through repeating this step for all folds and algorithms, are ranked from $1$ to $kn$, associated with the best result and the worst one, respectively~\citep{Garcia2010}. We call these new ranks ``aligned ranks''. The Friedman Aligned Ranks test statistic can be defined as}

\begin{equation}
    T ={\frac{(k-1)\Big[\sum_{j=1}^k \hat{R}_{.j}^2-(kn^2/ 4)(kn+1)^2\Big]}{\{[kn(kn+1)(2kn+1)]/6\}-(1/k)\sum_{i=1}^n\hat{R}_{i.}^2}}
\end{equation}
\textcolor{black}{where $\hat{R}_{.j}$ is the aligned rank total of the $j$th algorithms, $k$ is the number of algorithms, $\hat{R}_{i.}$ is the aligned rank total of the $i$th fold, and $n$ is the number of folds.} 

\textcolor{black}{If we reject the null hypothesis (i.e., the algorithms do not behave similarly and there is a significant difference between their performance), we need to apply a post hoc analysis to make a pairwise comparison. This pairwise comparison procedure detects which model performs better/worse than the others~\citep{Garcia2008, Garcia2010, Derrac2011}. We apply Bergmann-Hommel post hoc as it is the best-performing procedure recommended by~\citet{Derrac2011}. We use the \texttt{scmamp} package for performing both the Friedman Aligned Ranks test and Bergmann-Hommel post hoc analysis in the \texttt{R} environment.}

\subsection{\textcolor{black}{Spatial efficiency measure (SPAEF)}} 
\textcolor{black}{After sensitivity analysis of all algorithms in response to the transit improvement policy, we map the spatial distribution of newly generated transit trips estimated by all regressors. We aim to compare the spatial patterns obtained from the estimated transit trips by each algorithm. Hence, a multi-component metric suggested by~\citet{Demirel2018} is utilized}       

\begin{align}
    SPAEF &= 1-\sqrt{(\alpha-1)^2+(\beta-1)^2+(\gamma-1)^2} %\\
\end{align}
\textcolor{black}{where $\alpha$ is the Pearson correlation coefficient between map A and B (i.e., $\alpha=\rho(A,B)$), $\beta$ is the ratio of coefficient of variations illustrating spatial variability (i.e., $\beta=\Big(\frac{\sigma_{A}}{\mu_{A}}\Big)/\Big(\frac{\sigma_{B}}{\mu_{B}}\Big)$), $\gamma$ is the percentage of histogram intersection, and $n$ is the number of bins (i.e., $\gamma = \frac{\sum_{j=1}^n min(K_{j},L_{j})}{\sum_{j=1}^n K_{j}}$). For $\gamma$, the histogram $K$ of map A and the histogram $L$ of map B are computed. We followed the \texttt{Python} implementation of the SPAEF by the authors (\href{https://github.com/cuneyd/spaef}{github.com/cuneyd/spaef}) to implement it~\citep{Koch2018}.}

%%%%%%%%%%%%%%%%%%%%%%%%%%%%%%%%%%%%%%%%
%%%%%%%% Model interpretation %%%%%%%%%%
\subsection{\textcolor{black}{Model interpretability tools}}
In addition to maximizing the prediction performance of a model, exploring which features affect the prediction outcome is essential for creating new knowledge about travel behaviour. Statistical models are categorized as intrinsically interpretable algorithms in which the coefficients of models readily reveal the significance and direction of each feature's impact on output. Conversely, it is generally assumed that ML models are ``black-box'' since their prediction results cannot be interpreted directly by the model. Interpretability is one of the main concerns when it comes to adopting ML algorithms~\citep{Kim2020, Kim2021, Koushik2020}. However, several post hoc interpretability techniques have been recently developed but seldom used in ML applications within travel behaviour research. To understand how a model predicts and which variables and to what extent contribute to its prediction, global and local interpretability tools are applied. \textcolor{black}{We note that employing an intrinsically interpretable model is always preferred if the difference in its performance and that of black-box models is insignificant. }

\subsubsection{Global interpretability} 
Global interpretability clarifies how a model predicts in general and what is the entire behaviour of the model~\citep{Molnar2020}. \textcolor{black}{This approach quantifies the relationship and contribution of each feature to the model's prediction. Below are the descriptions for some of the interpretability tools.}
%%%%%%%%%%%%%%%%%%%%%%%%%%%%%%%%%%%%%%%%%
\paragraph{Feature importance} It is a widely used global interpretation technique calculated as the total effect of each feature on the final prediction. \textcolor{black}{It reflects how important a feature is for the predicted outcome of a model. The most used approach is permutation-based feature importance, in which the mean decrease in the performance of the out-of-bag sample is computed after permuting the values of a feature~\citep{Casalicchio2018, Breiman2001}. If the model's prediction error is increased after such permutation, it shows the feature is important, and the model's performance is sensitive to the change. On the other hand, shuffling the values of an unimportant feature does not significantly affect the model's performance. Ultimately, each feature is ranked by its variation in the model's prediction error after shuffling its values. You can refer to~\ref{sec:feature_importance_algorithm} for more details.}

%%%%%%%%%%%%%%%%%%%%%%%%%%%%%%%%%%%%%%%%%
\paragraph{Partial Dependence Plot (PDP)} It is a popular method to compute the partial relationship between one or a set of features and the targeted response~\citep{Friedman2001}. \textcolor{black}{This plot represents how changes in the distribution of one or two features affect the average expected outcome of the model while fixing the values of the remaining features. When there is no correlation between a feature and other predictors, the PDP accurately shows how the features of interest affect the average prediction. The computation of a PDP is straightforward to interpret; however, its main disadvantage is the independence assumption between attributes~\citep{Molnar2020}. The algorithmic way to obtain PDP is explained in~\ref{sec:PDP_algorithm}}.

\subsubsection{Local interpretability}
Another category of model-agnostic interpretation tools is the local interpretability technique emphasizing individual observations and examining the features' effects on the outcome per instance~\citep{Molnar2020}. ICE, SHAP, and LIME are examples of local interpretability tools.
%%%%%%%%%%%%%%%%%%%%%%%%%%%%%%%%%%%%%%%%%
\paragraph{Individual Conditional Expectation (ICE) plot} It is an extension of PDP that displays the effect of each attribute on the final prediction for individual observations \citep{Goldstein2015}. \textcolor{black}{Instead of calculating the average partial relationship, ICE plots represent how much a change in the value of a set of features affects the prediction of a single instance (See~\ref{sec:ICE_algorithm} for more details).}
\textcolor{black}{To generate both PDP and ICE plots, we use \texttt{scikit-learn} package (in \texttt{Python}).}

%%%%%%%%%%%%%%%%%%%%%%%%%%%%%%%%%%%%%%%%%
\paragraph{Shapley value (SHAP)} It is a recently developed model-agnostic technique whose values are defined as the unified measure of feature importance. SHAP values, computed based on cooperative game theory, represent the contribution of each feature to the final prediction of a specific instance~\citep{Lundberg2017, Molnar2020}. \textcolor{black}{This local contribution is measured by the difference between the prediction value of a specific observation per feature (independent variable) and the average prediction of a model according to various possible coalitions. When the feature $i$ joins a coalition $S$, its marginal contribution to the prediction $f_x$ is computed as} 
\begin{align}
    \Delta_i(x) = f_x(S)-f_x(S\setminus i).
\end{align}

\textcolor{black}{Therefore, the SHAP value evaluates the model's prediction and features' impact within every combination of features for each observation. The SHAP value ($\phi$) for feature $i$ at an observation $x$ using model $f$ can be calculated as}

\begin{align}
 \phi_i(f,x) = \sum_{S\subseteq({\{1,\dots,p\}}\setminus\{i\})} \frac{|S|!\;(p-|S|-1)!}{p!}\;\big(f_x(S)-f_x(S\setminus i)\big)
\end{align}
\textcolor{black}{where $p$ is the number of features, $S$ is a subset of features, $f_x(S)$ is the prediction value including all features in a subset of $S$ and $f_x(S\setminus i)$ is the prediction value of a subset $S$ without feature $i$. For this study, we use the \texttt{shap} package in \texttt{Python} to compute SHAP values.}

%%%%%%%%%%%%%%%%%%%%%%%%%%%%%%%%%%%%%%%%%
\paragraph{Local Interpretable Model-agnostic Explanations (LIME)} Like SHAP values, this method also provides local explanations of any model and shows the heterogeneity of individual observations \citep{Ribeiro2016}. These explanations are given by approximating the underlying black-box model to a simple interpretable model, e.g., linear models or decision trees, around a single input point \citep{Molnar2020}. LIME perturbs numeric data features using standard normal distribution and categorical features according to the training distribution. It then learns locally weighted linear models on the attribute-space neighbor data points of a specific observation. Accordingly, it locally interprets the predicted values of an observation through an interpretable model. \textcolor{black}{Thus, LIME models with the interpretability constraint explain the instance $x$ using the following notation}

\begin{align}
    \mathcal{L}(x) & = \text{arg}\min_{g\in G} {L(f, g, \pi_x) + \Omega(g)}
\end{align}
\textcolor{black}{where $g$ is an interpretable model (e.g., the linear regression model), which minimizes the loss $L$ (e.g., RMSE), and $\Omega$ is the model complexity (e.g., the number of features). Hence, we want to minimize the difference error between the original model $f$ and the explanation. On the other hand, $\pi_x$ determines the proximity range around instance $x$ that we consider for its explanation (Please refer to~\ref{sec:LIME interpretability}). 
We use the \texttt{lime} package in \texttt{Python} to implement the LIME interpretation.}

%%%%%%%%%%%%%%%%%%%%%%%%%%%%%%%%%%%
%%%%%%%%%%%% RESULTS %%%%%%%%%%%%%%
%%%%%%%%%%%%%%%%%%%%%%%%%%%%%%%%%%%
%\section{Experiments}
\section{Results}
\begin{figure}[!ht]
    \centering
    \includegraphics[width=0.85\textwidth]{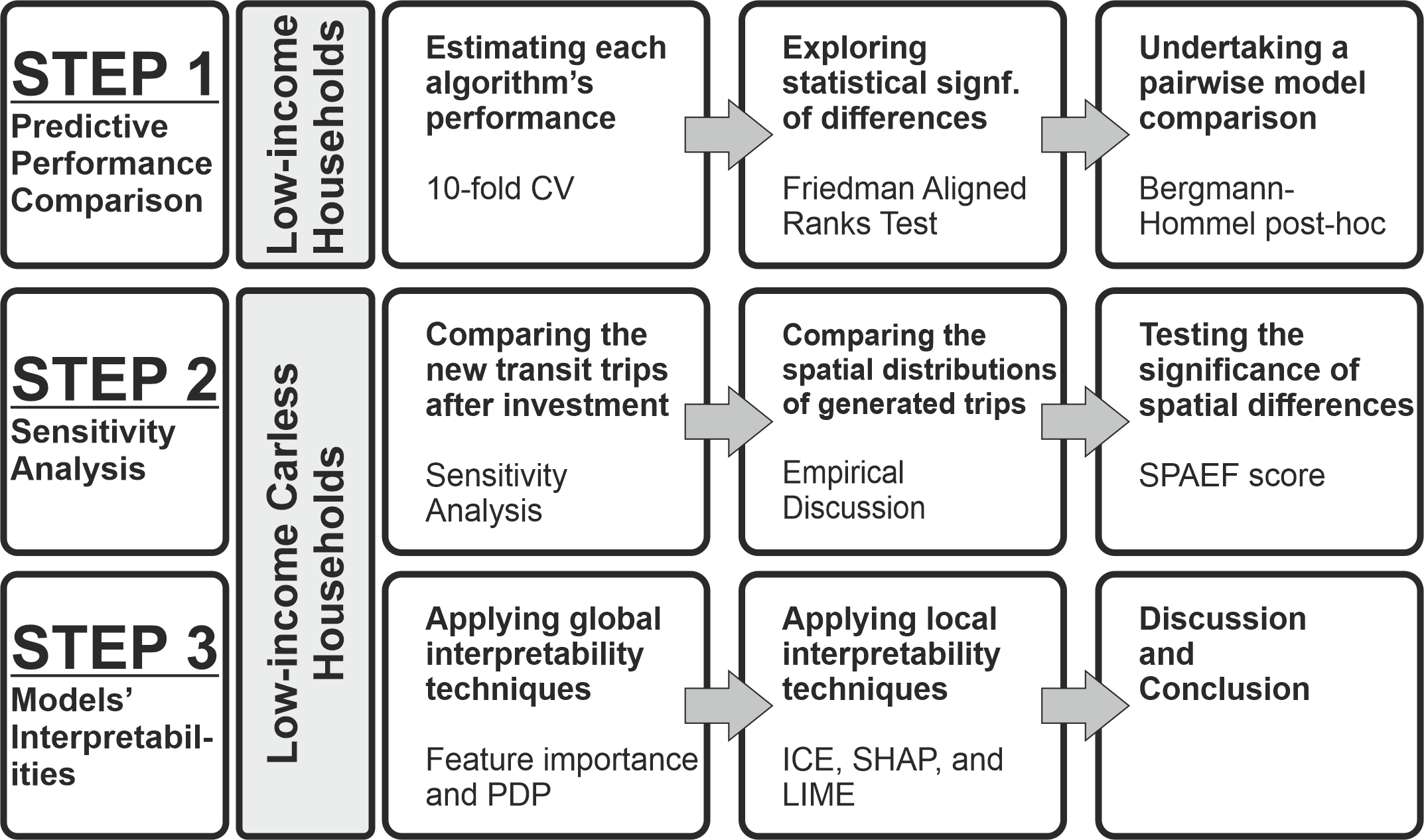}
    \caption{\textcolor{black}{The experimental design of the study}}
    \label{fig: flowchart}
\end{figure}

In this section, we present our experiment design and its results. 
\textcolor{black}{We first explain how to evaluate and compare algorithms' performance for low-income people.
The comparison is done to estimate the number of transit trips per person. Afterward, we explore the sensitivity of the classical and ML models to transit improvements in the low-income carless community since \citet{Allen2020} and \citet{Yousefzadeh2021} showed that they are more sensitive to transit improvements in this study area. Figure~\ref{fig: flowchart} illustrates a detailed diagram outlining the steps undertaken and the dataset used in this section.}

\subsection{Comparing \textcolor{black}{predictive performance of algorithms}}
%on estimating the number of transit trips}
To predict the number of transit trips, we utilize five ML regressors (DT, RF, XGB, NN, and SVM) and three statistical models as the baseline. In this step, we employ Hurdle and ZINB as count models in our comparison due to an excessive number of non-transit users in the dataset. LinR is another statistical method we used as the baseline. \textcolor{black}{The dataset used for this step consists of travel behaviour of low-income households. Each model's predictive performance is estimated using 10-fold cross-validation technique discussed in Section~\ref{CV}.} Table~\ref{table: regression} illustrates the average performance of each regressor across ten folds. Tree-based algorithms, i.e., RF and DT, demonstrate a smaller error in predicting the number of transit users. The results show that RF alongside NN and XGB has a higher R-squared value on average. Considering RMSE and RRSE, we arrive at the same conclusion; however, DT is the one that outperforms all other algorithms in terms of Median Absolute Error. \textcolor{black}{Since DT does not demonstrate a high performance in terms of other metrics, we may conclude that its median error is low (i.e., it is able to predict some observations better than other algorithms), while it has many outliers in its prediction error (i.e., it cannot have a proper prediction for some observations, which causes a higher RMSE). This observation might be inferred as an instability in prediction plus an overfitting case with a high variance and low bias. Therefore, we still prefer RF to DT as it is able to alleviate overfitting issues without the need for pruning~\citep{ZHOU2020106931}.}

\begin{table}[!ht]
    \centering
    \caption{Performance comparison of each regressor for predicting the number of transit trips in individuals' daily trips based on 10-fold cross-validation}
    \resizebox{0.99\textwidth}{!}{
    \begin{tabular}{lc|rrrrrrrr}
        \toprule
        \multicolumn{10}{l}{\textcolor{black}{\textit{Dependent Variable = The number of transit trips}}} \\
        \midrule
        \textbf{} & \textbf{} & \textbf{DT} & \textbf{RF} & \textbf{XGB} & \textbf{NN} & \textbf{SVM}  & \textbf{LinR}  & \textbf{ZINB} & \textbf{Hurdle} \\
        \midrule
        \multicolumn{1}{l|}{\multirow{2}{*}{\textbf{R-Squared \scriptsize{(\%)}}}}     & \textbf{$\bar{X}$} & 52.33 & \textbf{58.22} & 53.90 & 55.37 & 53.06 & 46.74 & 49.89 & 48.11 \\
        \multicolumn{1}{l|}{}  & \textbf{$s$}& 1.84 & 1.10 & 1.42 & 1.50 & 1.55 & 1.61 & 2.02 & 2.16 \\
        \multicolumn{1}{l|}{\multirow{2}{*}{\textbf{RMSE loss \scriptsize{(\%)}}}}     & \textbf{$\bar{X}$} & 69.46 & \textbf{65.04} & 68.31 & 67.22 & 68.93 & 73.43 & 71.21 & 72.46 \\
        \multicolumn{1}{l|}{}   & \textbf{$s$}& 1.30 & 1.43 & 1.46 & 1.64 & 1.67 & 1.63 & 1.48 & 1.62 \\
        \multicolumn{1}{l|}{\multirow{2}{*}{\textbf{MedAE loss\scriptsize{(\%)}}}}     & \textbf{$\bar{X}$} & \textbf{6.82} & 11.60 & 12.14 & 9.86 & 8.09 & 20.33 & 10.97 & 14.68 \\
        \multicolumn{1}{l|}{}   & \textbf{$s$} & 0.26 & 0.90 & 0.68 & 1.46 & 0.21 & 0.84 & 0.72 & 0.55 \\
        \multicolumn{1}{l|}{\multirow{2}{*}{\textbf{RRSE loss\scriptsize{(\%)}}}}     & \textbf{$\bar{X}$} & 69.03 & \textbf{64.63} & 67.89 & 66.80 & 68.50 & 72.97 & 70.77 & 72.02 \\
        \multicolumn{1}{l|}{}   & \textbf{$s$}& 1.33 & 0.86 & 1.05 & 1.13 & 1.13 & 1.11 & 1.43 & 1.50 \\

        \bottomrule
    \end{tabular}
}
    \label{table: regression}
\end{table}

To investigate the significance of performance differences, we employ statistical tests. Accordingly, we select multiple statistical comparison tests for the folds~\citep{Garcia2010}. We choose Friedman Aligned Ranks as a non-parametric test since the sample size is small. The null hypothesis is that there is no significant difference in the algorithms' performance. The $p$-value of the test for the obtained RMSE measure is $1.08e-01$ given ten folds. Hence, we conclude that differences in algorithms' performances are statistically significant (for $\alpha = 0.05$). We apply the same test for all metrics, and a similar conclusion is reached. 

\begin{figure}[!ht]
    \centering
    \begin{subfigure}{.45\textwidth}
        \centering
        \includegraphics[width=\textwidth]{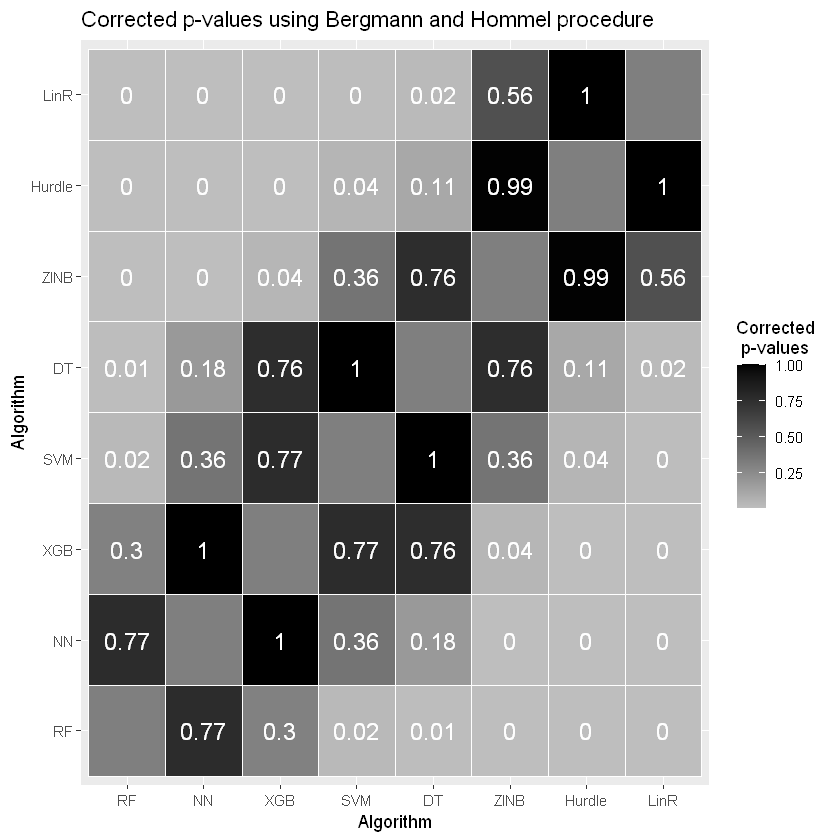}
        \caption{Corrected    pairwise    $p$-values using Bergmann-Hommel post hoc procedure}
        \label{fig: FTest_reg}
     \end{subfigure}
     \hspace{0.7cm}
     \begin{subfigure}{.45\textwidth}
        \centering
         \includegraphics[width=0.4\textwidth]{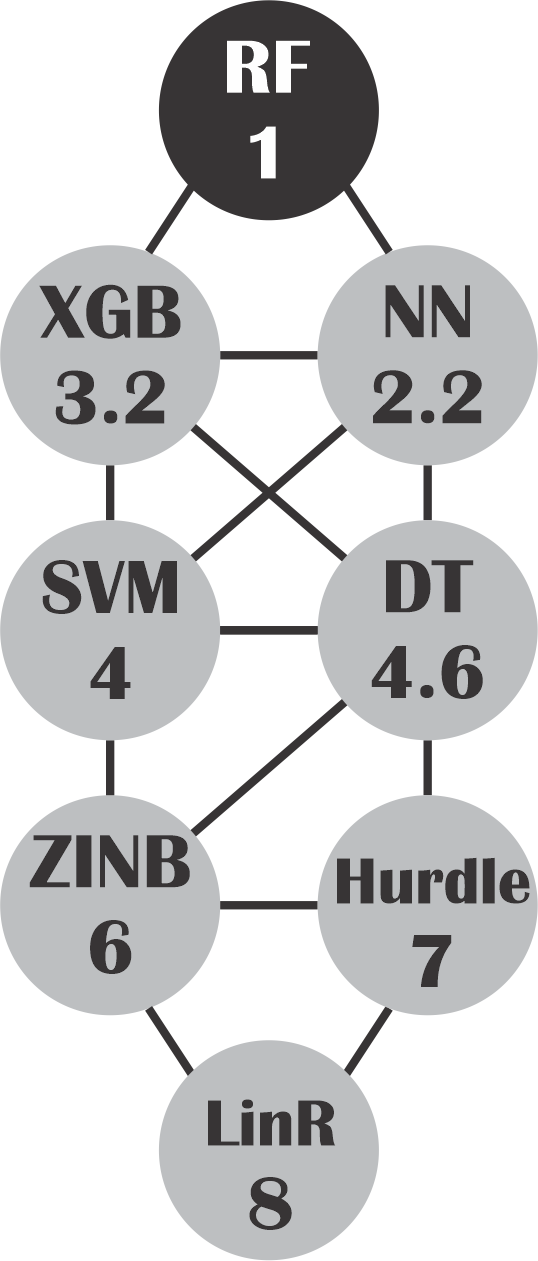}
         \caption{Average rank of regressors (${\alpha=0.05}$). Edges between algorithms indicate an insignificant difference.}
         \label{fig: FTest_reg_graph}
    \end{subfigure}
    \caption{Friedman Aligned Ranks test result of regression ``RMSE'' values after Bergmann-Hommel post hoc procedure}
    \label{fig: FTest_regression}
\end{figure}

To explore the source of the difference, we apply a post hoc analysis. The post hoc procedure assesses the difference between all algorithms in terms of the absolute difference in the average ranking. It enables us to have a pairwise comparison among models. Moreover, adjusted $p$-values are computed using the post hoc procedure. Figure~\ref{fig: FTest_reg} is the matrix of corrected $p$-values for RMSE measure after applying the Bergmann-Hommel post hoc analysis. Dark colors show higher adjusted $p$-values, representing an insignificant difference between pairs of algorithms in terms of the RMSE metric. For instance, according to the Bergmann-Hommel post hoc test, there is no evidence for a statistical difference between RF and XGB in terms of RMSE values. On the other hand, Figure~\ref{fig: FTest_reg_graph} shows only the insignificant pairwise difference among algorithms. The number on each node indicates the average rank of the algorithm given ten folds. Accordingly, RF has the lowest average rank of 1 and is on a par with NN and XGB in terms of RMSE. More details on the differences between ML models and the traditional baseline for other metrics are provided in \ref{sec:detailed_comp}.
 
\subsection{Sensitivity analysis}
\begin{figure}[!ht]
    \centering
    \includegraphics[width=0.75\textwidth]{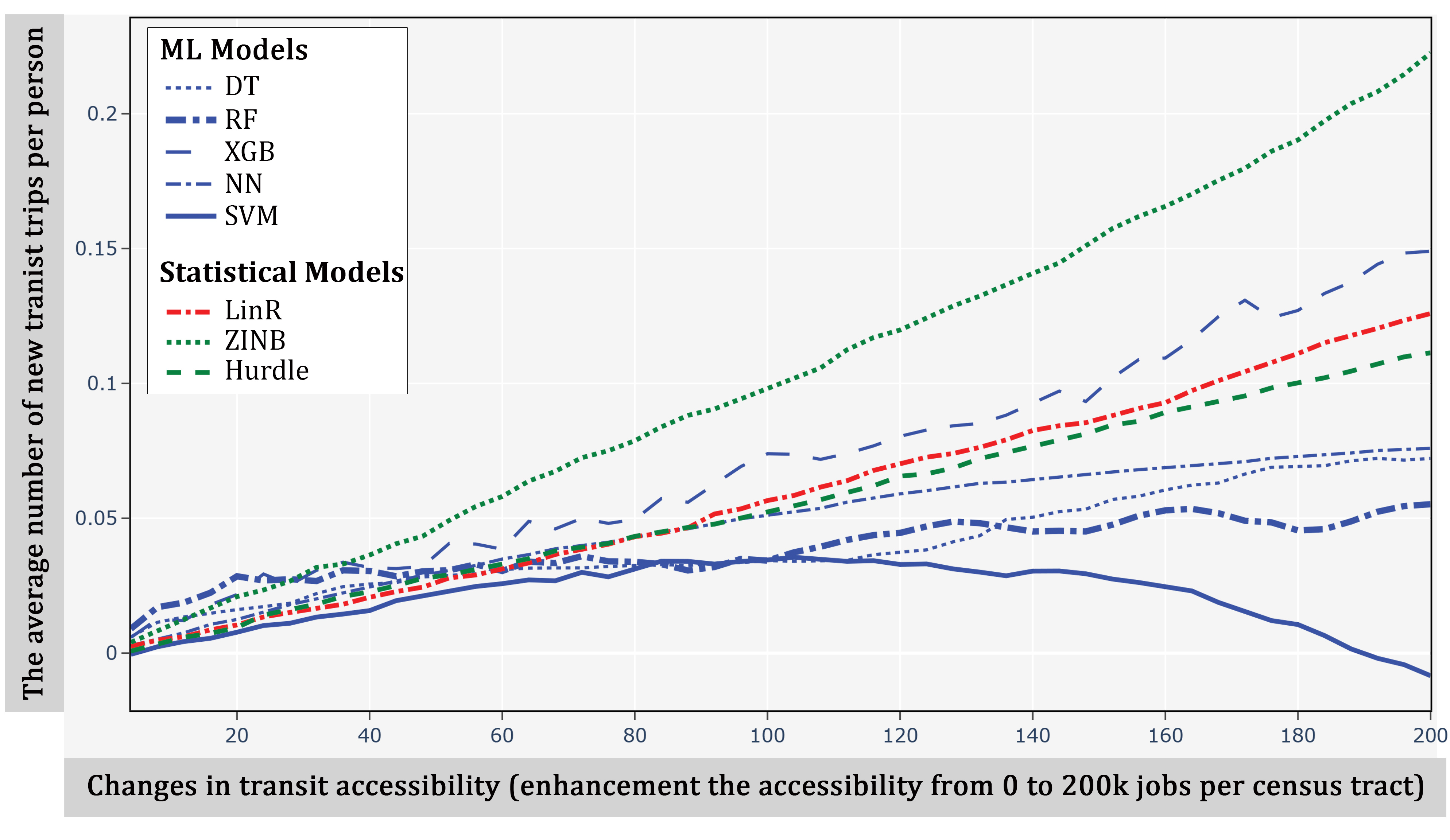}
    \caption{\textcolor{black}{The sensitivity of all models to the accessibility improvement for low-income carless households (The baseline model is shown in red, the statistical models in green, and the ML models in blue.)}}
    \label{fig: sensitivity_all}
\end{figure}

To evaluate transit investment policy in low-income communities, we investigate the effect of transit accessibility improvements on taking transit using sensitivity analysis of all regressors. Figure~\ref{fig: sensitivity_all} shows the average number of new transit trips after incrementally increasing accessibility for low-income zero-car individuals. \textcolor{black}{We filtered our dataset on low-income carless households since they are shown to be more sensitive to transit improvements~\citep{Allen2020, Yousefzadeh2021}. We select the LinR model as our baseline model, coloured red in Figure~\ref{fig: sensitivity_all}. Since it is an interpretable, simple, and widely used model in the literature, it is considered a reference model. Accordingly, we compare its predictions with those of other models.} We focus on low-income carless households as a subset of our dataset to compare the predictive outcomes of all models for the disadvantaged group. Among all regressors, the ZINB model predicts the average number of new transit trips after increasing accessibility to 200k jobs by more than 0.2 per person. Comparing the variations in the predicted trip numbers using statistical models represents that all ZINB, LinR, and Hurdle models follow a roughly identical trend. \textcolor{black}{However, the average number of predicted transit trips after increasing accessibility to 200k jobs using LinR and Hurdle models is half of the ZINB ones.} \textcolor{black}{Comparing the results of the SVM algorithm with those of the reference model, we can see that} the curve of newly generated transit trips using the SVM model increases up to 100k jobs. Then, it starts decreasing until it reaches 0 in 200k jobs. Such behaviour is also not observed by other models. Moreover, the predictive behaviour of ensemble models like RF and XGB is similar up until 60k jobs, but their prediction lines diverge beyond that. \textcolor{black}{Disregarding the special case of XGB, all statistical models act as an upper bound for the ML models, claiming a significantly higher sensitivity of low-income carless households to transit investment.} We conclude that there is a difference among all models in the average number of new transit trips after the largest accessibility gain (i.e., 200k jobs). Therefore, model selection may impact policy evaluation. To choose the optimal model, a researcher may consider the predictive performance and the interpretation of features contributing to that prediction. Before discussing possible interpretations of the best-performing model, we aim to explore how these prediction differences are spatially distributed in the region.

\begin{figure}[!ht]
    \centering
    \includegraphics[width=0.95\textwidth]{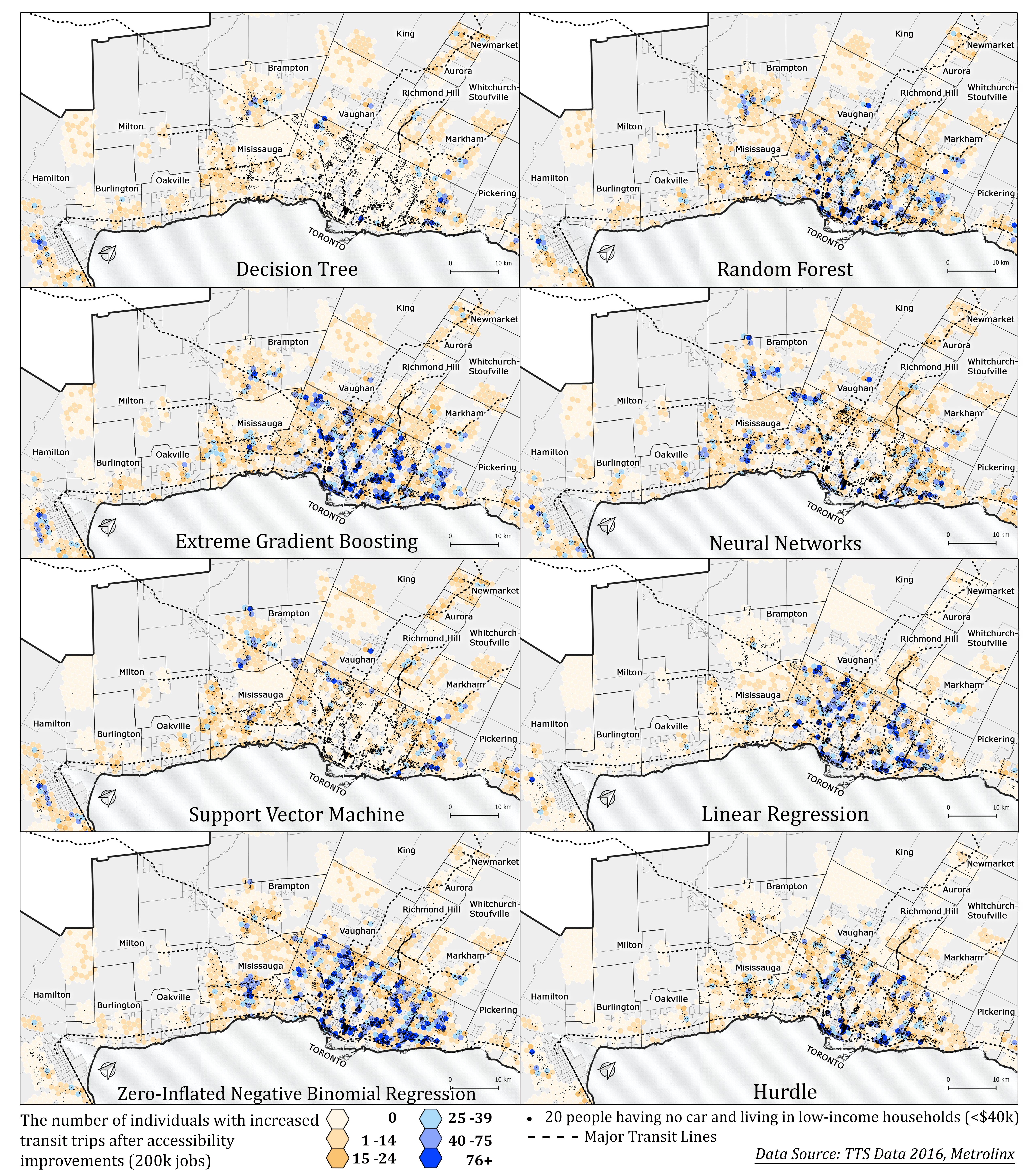}
    \caption{\textcolor{black}{The spatial prediction of newly generated transit trips by low-income carless group after improving job accessibility (enhancement the accessibility from 0 to 200k jobs per census tract) using different algorithms}}
    \label{fig: sensitivity}
\end{figure}

\textcolor{black}{We mapped the spatial distribution of newly generated transit trips by low-income carless households after increasing the transit accessibility level by 200,000 jobs (See Figure~\ref{fig: sensitivity}). Such accessibility gain can be achieved by investments in higher-order transit services, as discussed by~\citet{Farber2017}. We created $1000\times1000\text{m}$ hexagonal maps to investigate spatial similarities/differences between models' predictions of new transit ridership. We observed a clear visual distinction in the spatial distribution of new transit trips predicted by each model in the whole study area (GTHA), particularly in the inner suburbs and downtown of Toronto, the city of Hamilton, Brampton, and Newmarket regions. Our maps showed that the statistical models have small numbers of new transit trips in the Hamilton, Brampton, and Newmarket regions. However, ML models suggest new transit trips after accessibility improvement in the same areas. In a special case of SVM, all the new transit trips belong to the inner suburb of Toronto, and we observe a negligible number of transit trips in Downtown Toronto. This observation is consistent with Figure~\ref{fig: sensitivity_all} in which SVM shows a decline in the number of new transit trips if the accessibility is significantly improved. Based on both Figures~\ref{fig: sensitivity_all} and \ref{fig: sensitivity}, only XGB has a similar number of transit trips and spatial patterns to those of the statistical methods.} 

\begin{figure}[!ht]
    \centering
    \includegraphics[width=0.65\textwidth]{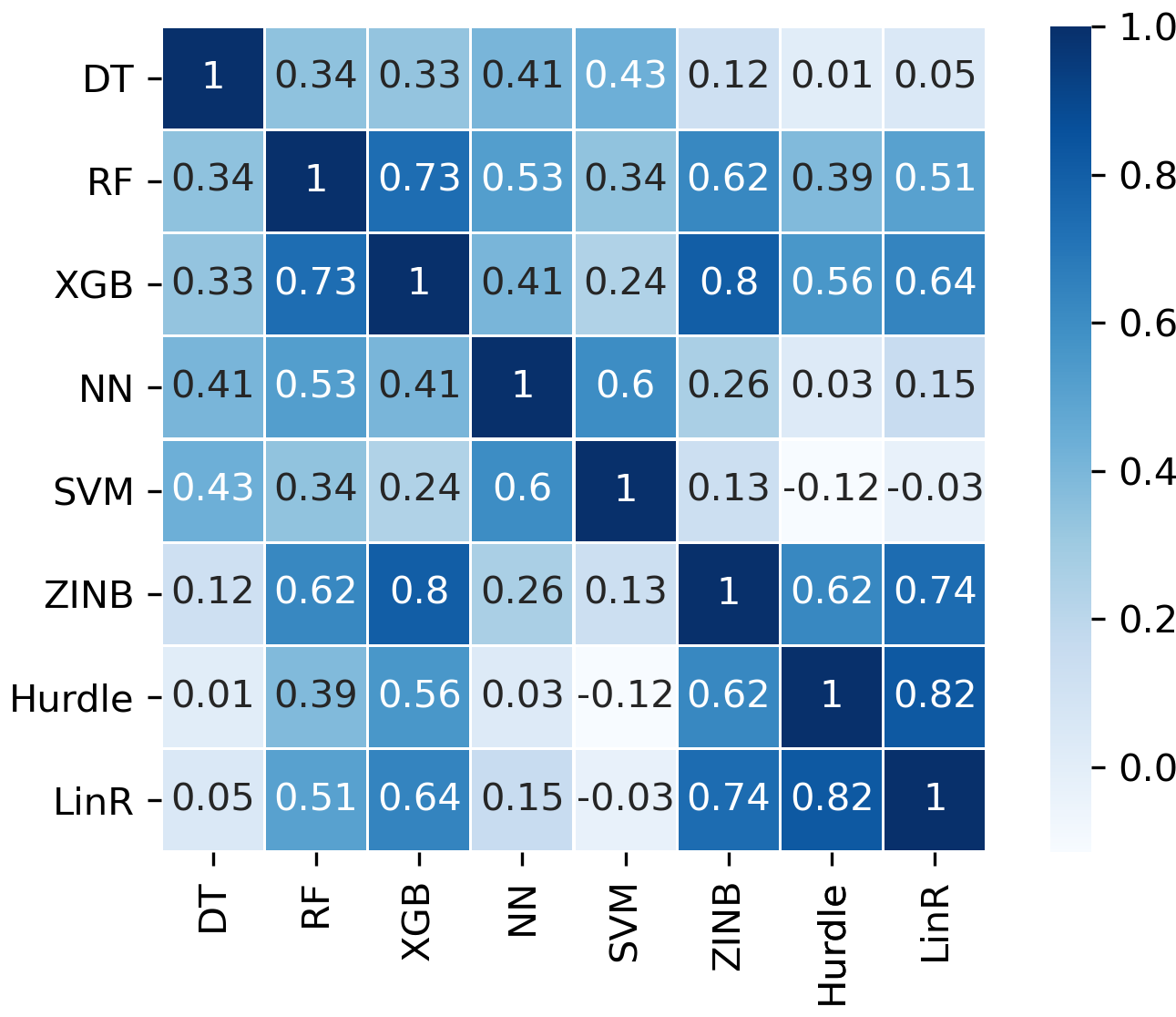}
    \caption{The heatmap of SPAEF scores}
    \label{fig: SPAEF}
\end{figure}

Besides the visual evaluation, we also aim to statistically compare whether there is a significant difference between the spatial patterns of all maps. For this reason, we apply the SPAtial EFficiency metric (SPAEF) \citep{Demirel2018}. This metric considers three statistical measures, including Pearson correlation, coefficient of variation, and histogram overlap, and their outputs are integrated into one measure. SPAEF values calculated for each map are reported in Figure~\ref{fig: SPAEF}. The high scores of SPAEF at the right bottom of this heatmap illustrate that all three statistical models have high spatial similarities. On the other hand, they show a different spatial distribution than that of NN, DT, and SVM. Discarding the similarity of RF and XGB to traditional approaches, we can see a dark $5\times5$ cluster of ML models at the top left and another dark $3\times3$ cluster of traditional models at the bottom right. Thus, we conclude that utilizing ML algorithms instead of traditional models may suggest a different spatial pattern of transit use after accessibility improvements (see Figure~\ref{fig: sensitivity_all}) and may result in a different spatial policy recommendation at the end (see Figure~\ref{fig: sensitivity}).

\subsection{\textcolor{black}{Model interpretability}}
\textcolor{black}{In this section, we only consider the best performing ML model of this study, i.e., RF.} The interpretability approaches we use are model-agnostic; therefore, this process can be generalized to any other ML or even traditional model. We consider the number of transit trips per person as our dependent variable.

\begin{figure}
    \centering
    \includegraphics[width=0.85\textwidth]{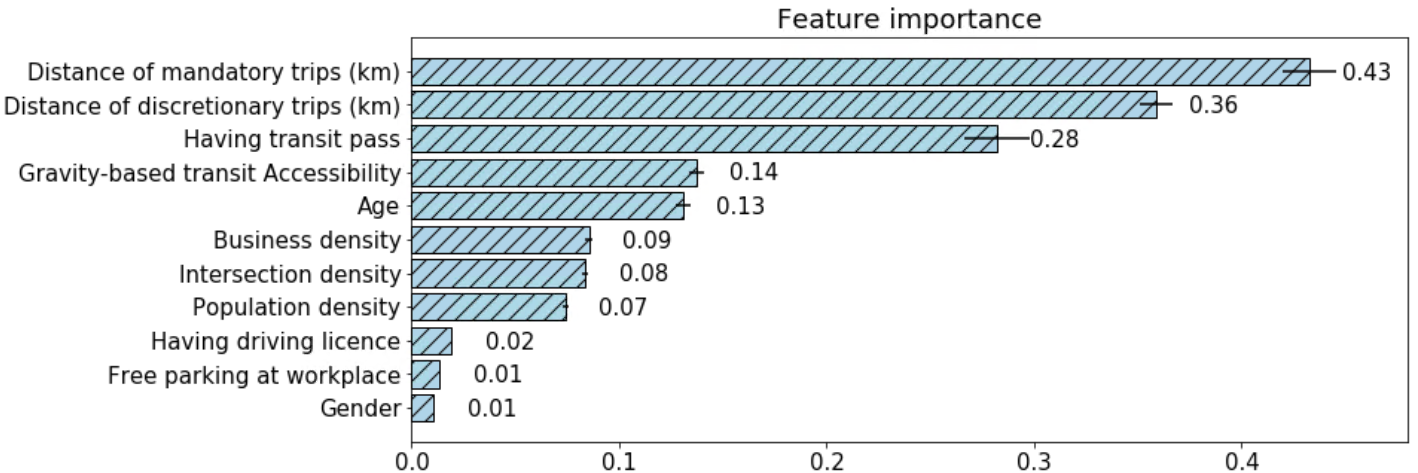}
    \caption{The importance of each feature for predicting the number of transit trips using RF model}
    \label{fig:feat_imp_RF}
\end{figure}

\textcolor{black}{Given the feature importance interpretation technique, we compute the total effect of each variable on the final outcome. Accordingly, Figure~\ref{fig:feat_imp_RF} shows} the influence of each independent variable on having transit trips for the low-income carless stratum. It denotes that the most significant variable in predicting the number of transit trips is the mandatory trip length. Also, transit accessibility is among the most important variables confirming that this measure is strongly associated with activity participation. However, it does not show whether this feature affects the output positively or negatively. We observe that driving licence possession, the free parking spot at the workplace, and gender have the lowest score, i.e., the least importance in transit trip prediction. It is aligned with our expectations of the variables. For example, the driving licence and free parking are immaterial if people do not own cars.

\begin{figure}
    \centering
    \includegraphics[width=\textwidth]{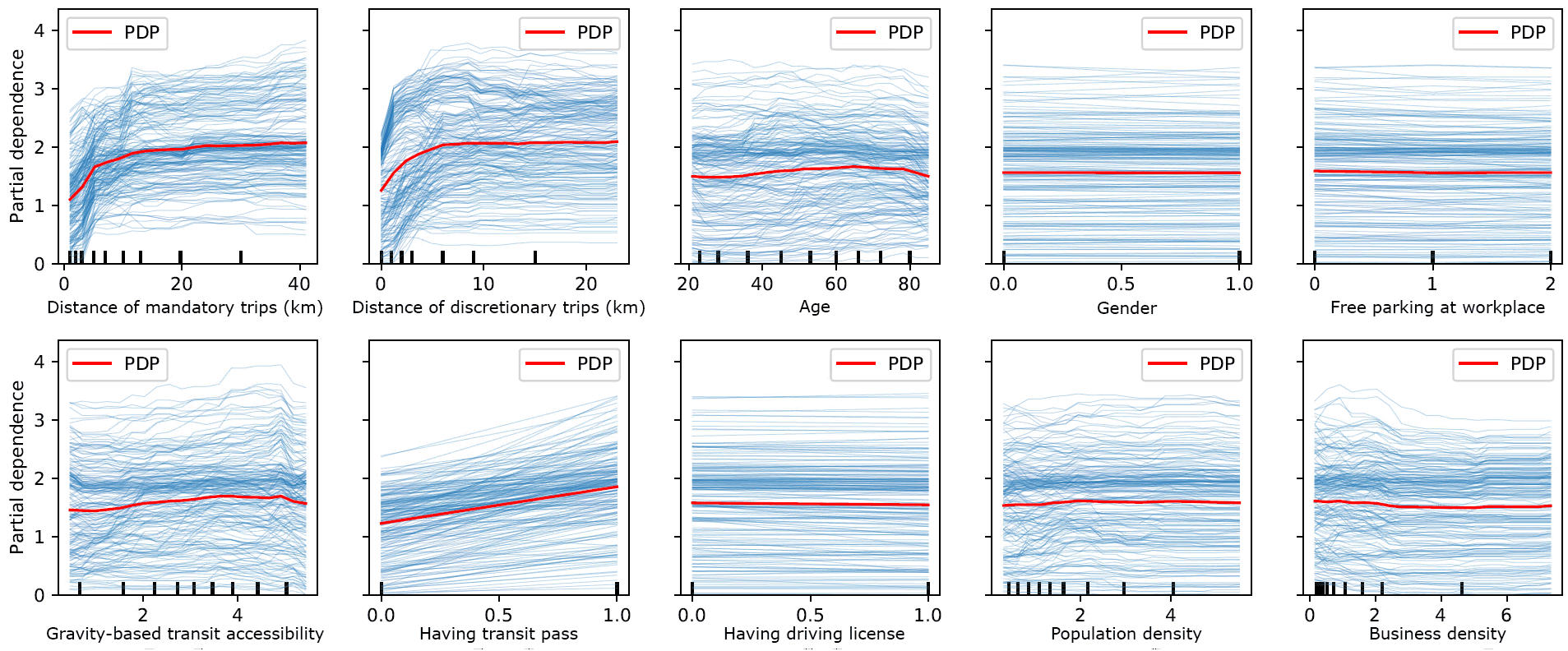}
    \caption{ICE and PD plots for predicting the number of transit trips after increasing transit accessibility by 200k jobs (Blue lines indicate ICE plots.)}
    \label{fig: ICE-PDP}
\end{figure}

\textcolor{black}{To determine the partial link between each attribute and the targeted variable, we draw PDP. Furthermore, we use ICE plot to understand the impact of each feature on the final prediction for individual instance. PD plot (red lines) in Figure~\ref{fig: ICE-PDP} illustrates} how the variations of each independent variable affect the expected average number of transit trips for vulnerable individuals. It also shows the direction of the effect between independent features and the dependent variable. The disadvantage of this technique is that it assumes there is no correlation between independent variables, whereas this assumption is often inaccurate in the real world. For instance, there might be some correlation between the number of people living in a region and its business density. However, PDP fails to show the mutual impact of these two variables. This plot indicates that the number of transit trips increases as either mandatory or discretionary trip lengths increase. However, it shows that gender, free parking at the destination, and driving licence do not influence one's number of transit trips.

\textcolor{black}{We also apply ICE as a local interpretation technique to see the effect of each variable on the outcome of each observation.} Each ICE line (blue lines) in Figure~\ref{fig: ICE-PDP} represents how the dependent variable changes when an independent feature changes for observation, while the PDP line defines the average of the line of an ICE plot \citep{Molnar2020}. This change in the dependent variable, e.g., the number of trips, is estimated by keeping other attributes intact and incrementally increasing the specific feature for a single observation, e.g., a single person. The plot shows 200 random individual observations within the dataset and depicts how the prediction of the number of transit trips changes as the independent variables change (blue lines). Unlike the PD plot showing the average effect of an independent variable on the output, we can see possible anomalies in the ICE plot. For instance, although the longer mandatory trips are, the more transit trips are taken, we observe some individuals for whom increasing the length of trips decreases their transit use. It is expected for suburbs with lower access to transit to use their own vehicle when their trip length is high. 
These individual-level findings cannot be obtained by merely checking the coefficients of a statistical model. We note that these local interpretation tools are generalizable to any other statistical or ML algorithm, facilitating the interpretability of any model with higher granularity.

\begin{figure}
    \centering
    \includegraphics[width=0.9\textwidth]{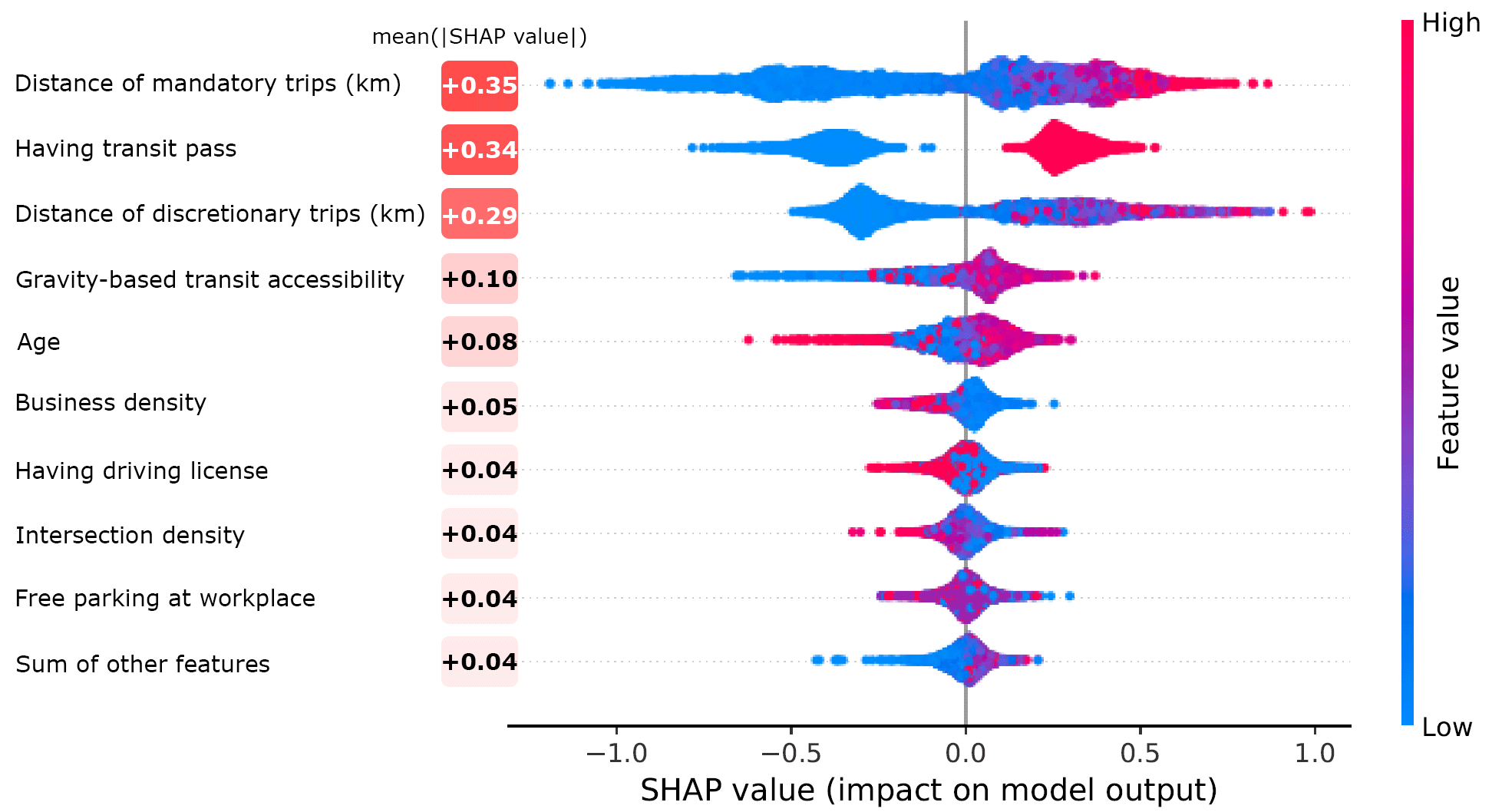}
    \caption{SHAP values' distribution and mean. Features are sorted by their mean SHAP values.}
    \label{fig:SHAP}
\end{figure}

\textcolor{black}{We also applied SHAP to see how each feature affects the prediction of a single instance in different coalitions (see Figure~\ref{fig:SHAP}).} In our study, Shapley values show the average contribution of each feature to the predicted number of transit trips across all possible coalitions of features, including and excluding this feature value. Therefore, it is useful when the contributions of features are unequal, but they may affect each other. The sum of Shapley values for all attributes of an individual is equal to the predicted number of trips for oneself subtracted from the mean predicted number of trips for everyone.

In Figure~\ref{fig:SHAP}, we observe that the length of the trips, whether mandatory or discretionary, together with having a transit pass contribute the most to the number of transit trips for low-income carless people -- i.e., they have the highest mean SHAP value. The density of the length of the mandatory and discretionary trips shows how common different trip lengths are in the dataset, and the coloring indicates a smooth increase in the log odds ratio of the transit use as the trip length increases. For the transit pass possession, unsurprisingly, we observe two clear clusters: people owning a transit pass have a higher number of trips and vice versa. A longer tail to the left for transit accessibility means that living in low transit-accessible regions, e.g., suburban, can significantly reduce the number of transit use, but high accessibility does not necessarily significantly raise the number of transit trips either. For instance, in downtown, where biking and walking to destinations are convenient and at the same time transit is accessible, low-income individuals may prefer active transportation.

\begin{figure}[!ht]
    \begin{subfigure}{.5\textwidth}
        \centering
        \includegraphics[width=1\textwidth]{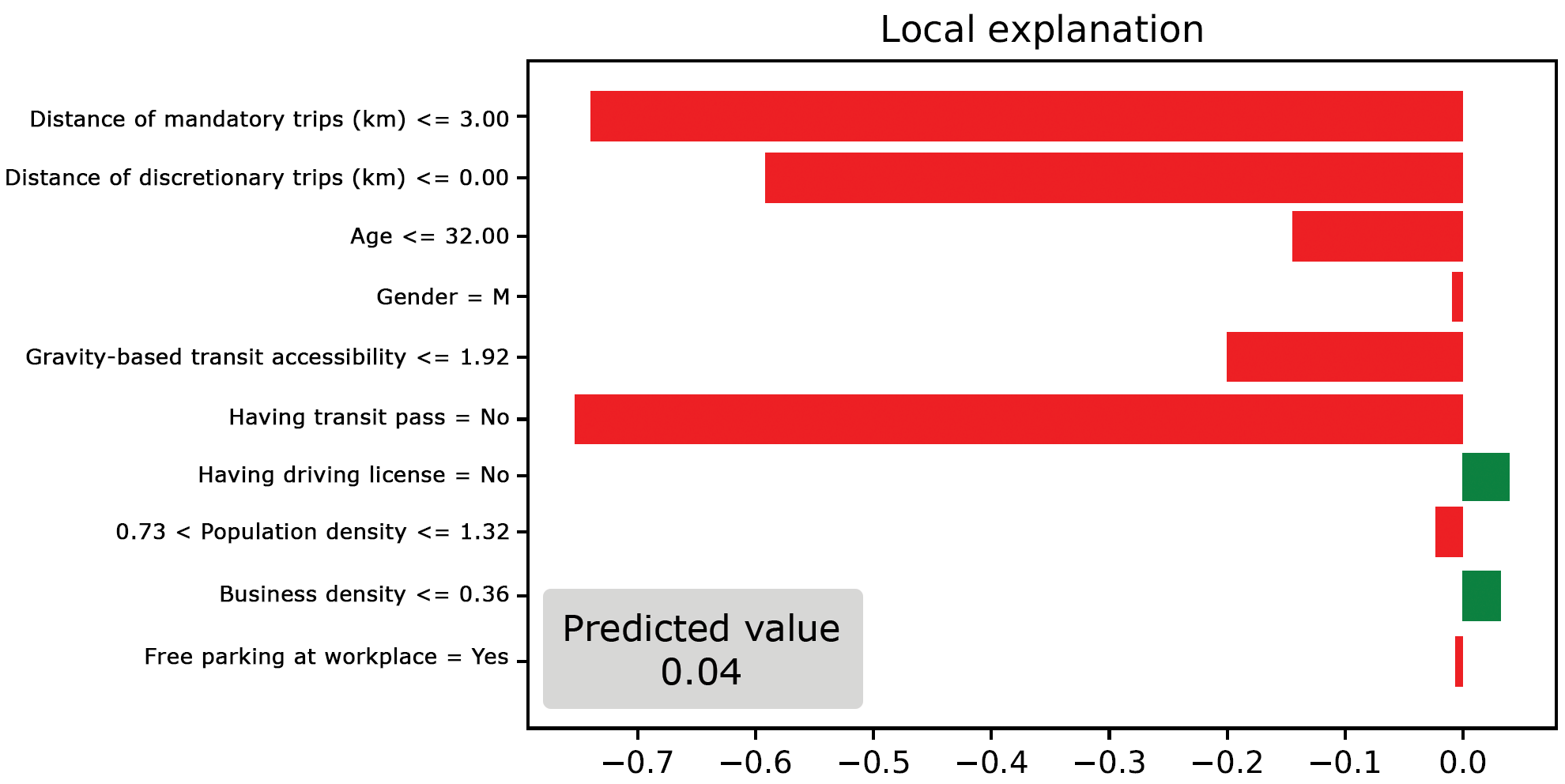}
        \caption{A non-transit-user individual.}
        \label{fig: LIME-non-user}
    \end{subfigure}
    ~
    \begin{subfigure}{.5\textwidth}
        \centering
         \includegraphics[width=1\textwidth]{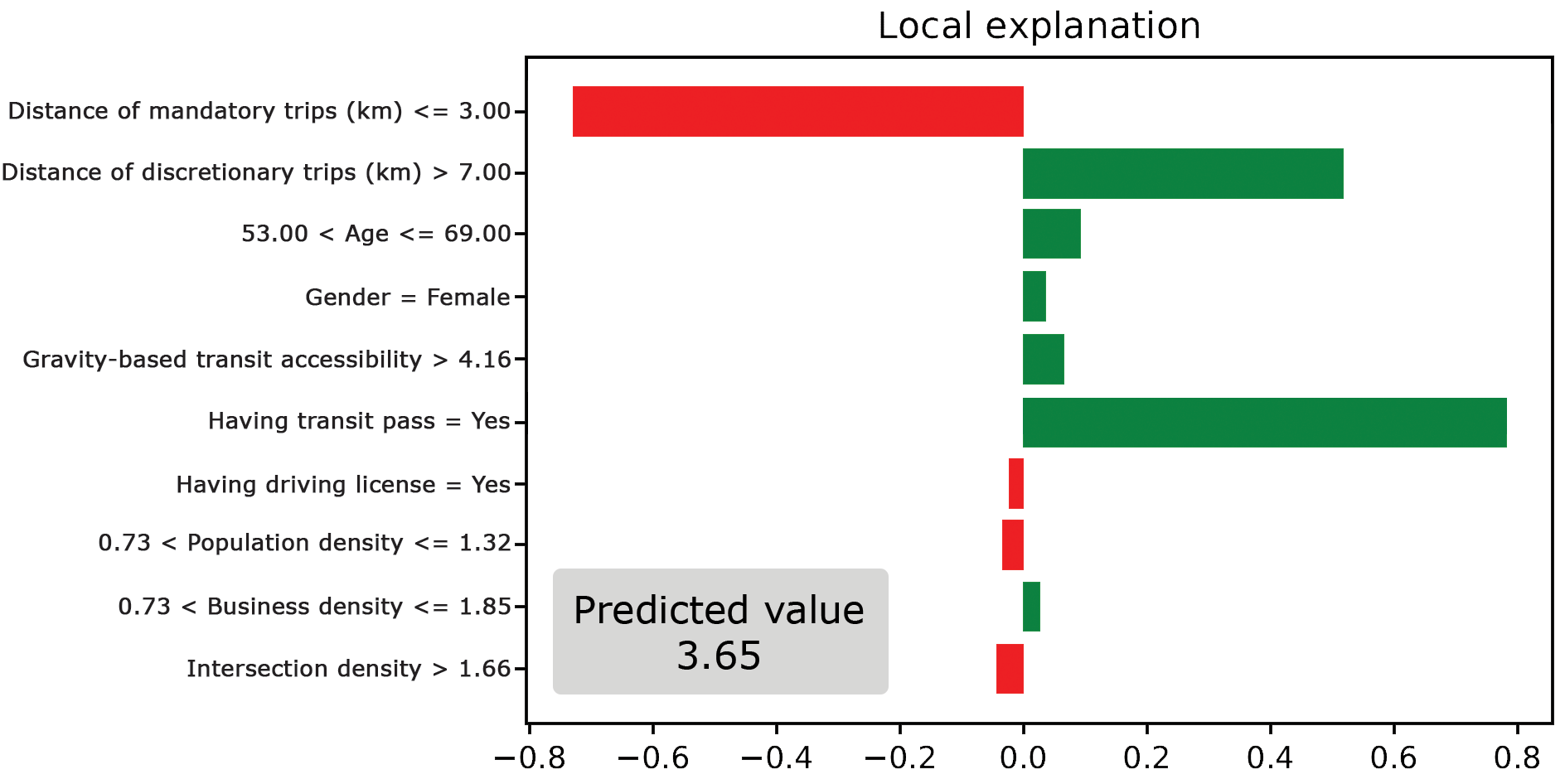}
         \caption{A transit-user individual.}
         \label{fig: LIME-user}
    \end{subfigure}
    \caption{Two sampled individuals, one not using transit and one a frequent transit user.} \label{fig: LIME-all}
\end{figure}

\textcolor{black}{To better understand the predicted values of a specific individual, we use the LIME tool.} Figure~\ref{fig: LIME-all} shows two randomly selected individuals, one who does not use transit and one who has five transit trips a day. LIME can be used to explore the notion behind the predicted values for a specific user. As the author of the original paper mentioned, it is also a way to check a model's trustability~\citep{Ribeiro2016}. We consider two extreme cases and see how each feature contributes to the final prediction. The $y$-axis shows the condition that holds for the feature value, and the $x$-axis shows the feature effect, i.e., its weight times its actual value. Figure~\ref{fig: LIME-non-user} for the non-transit user shows that not having a transit pass, having short trips, being middle age, and having low transit access lead to his preference for other travel modes. Figure~\ref{fig: LIME-user} for the frequent transit user indicates that her high number of discretionary trips, better transit accessibility, having a transit pass, and being older have an impact on her frequent transit trips. These detailed observations per individual are only possible when we use local interpretability. Therefore, besides the global interpretability of the models, a higher granularity of the interpretation sheds light on the model decisions and the soundness of its predictions.

\subsection{\textcolor{black}{Lessons for equity analysis}}
\begin{figure}[!ht]
    \centering
    \includegraphics[width=0.65\textwidth]{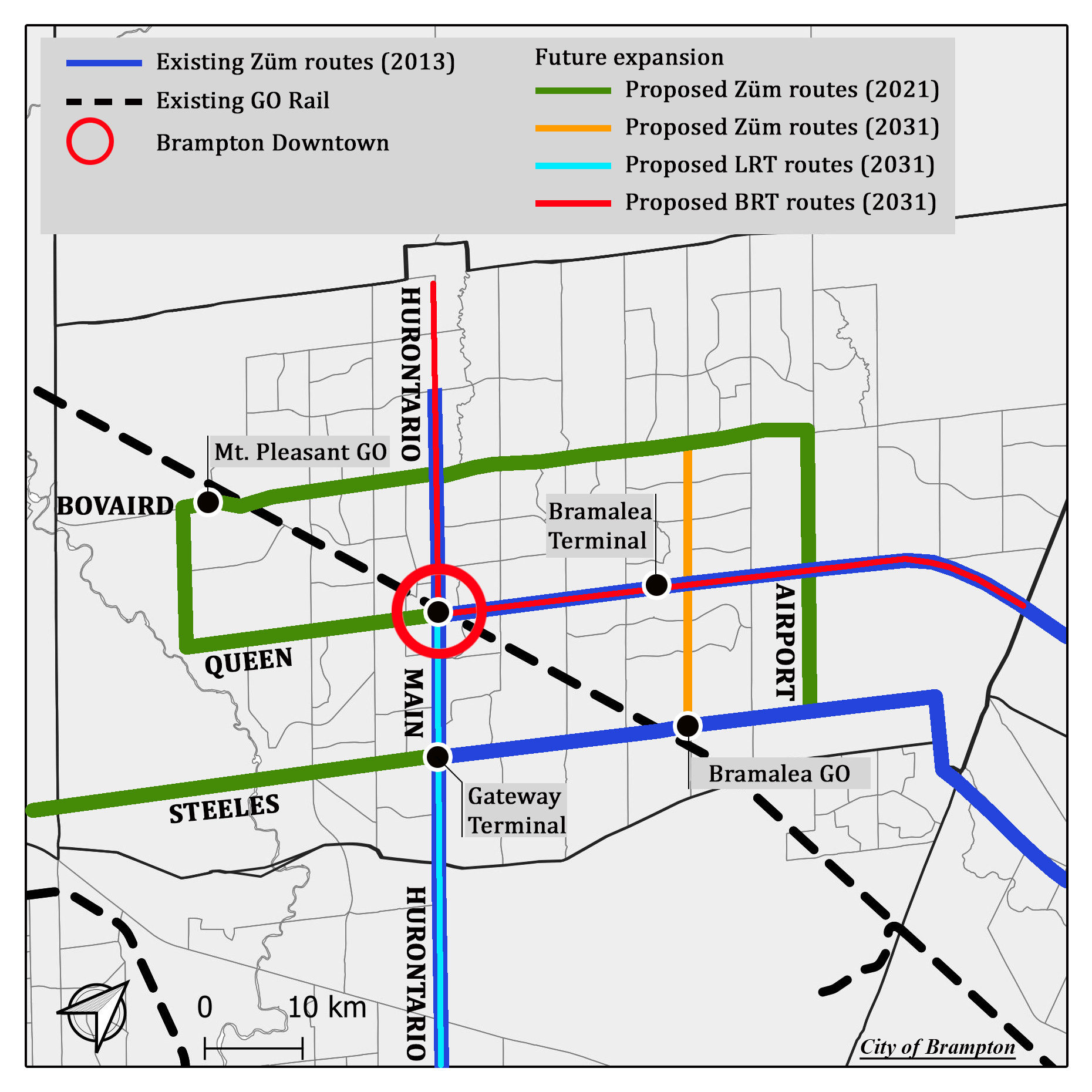}
    \caption{\textcolor{black}{Proposed future transit network for the city of Brampton~\citep{Barmpton2015}}}
    \label{fig: Brampton_plan}
\end{figure}

\textcolor{black}{To better investigate the spatial difference in models' predictions for the low-income carless community, we mainly focus on the city of Brampton. Figure~\ref{fig: Brampton_plan} shows the existing network of Brampton in 2016 and its recommended rapid transit and transportation network by 2031~\citep{Barmpton2015}. Low-income households are mainly located in Brampton central~\citep{Allen2020}, hosting connections between regional and local transit networks. Northern Brampton, on the other hand, is the least serviced region in the city, where still some transit-dependent low-income households are living~\citep{gullusci2021Brampton}.  The city planned to extend the Züm BRT corridor\footnote{Züm BRT Routes are higher-order transit routes that are intended to offer regular, high-capacity and high-quality service. Züm BRT Routes provide better and faster connectivity between people and locations by deploying innovative vehicles equipped with intelligent technology systems.} to Queen West Street and Bovaird Drive by 2021; however, due to the COVID-19 pandemic, it is only partially implemented by 2022. The existing Züm facility mainly operates in mixed traffic, whereas the city plans to dedicate the exclusive lanes for the BRT in Northern Hurontario and Queen East streets by 2031 (See the red line in Figure~\ref{fig: Brampton_plan})~\citep{Barmpton2015}. This extension makes northern Brampton more accessible. }

\begin{figure}[!ht]
    \centering
    \includegraphics[width=0.95\textwidth]{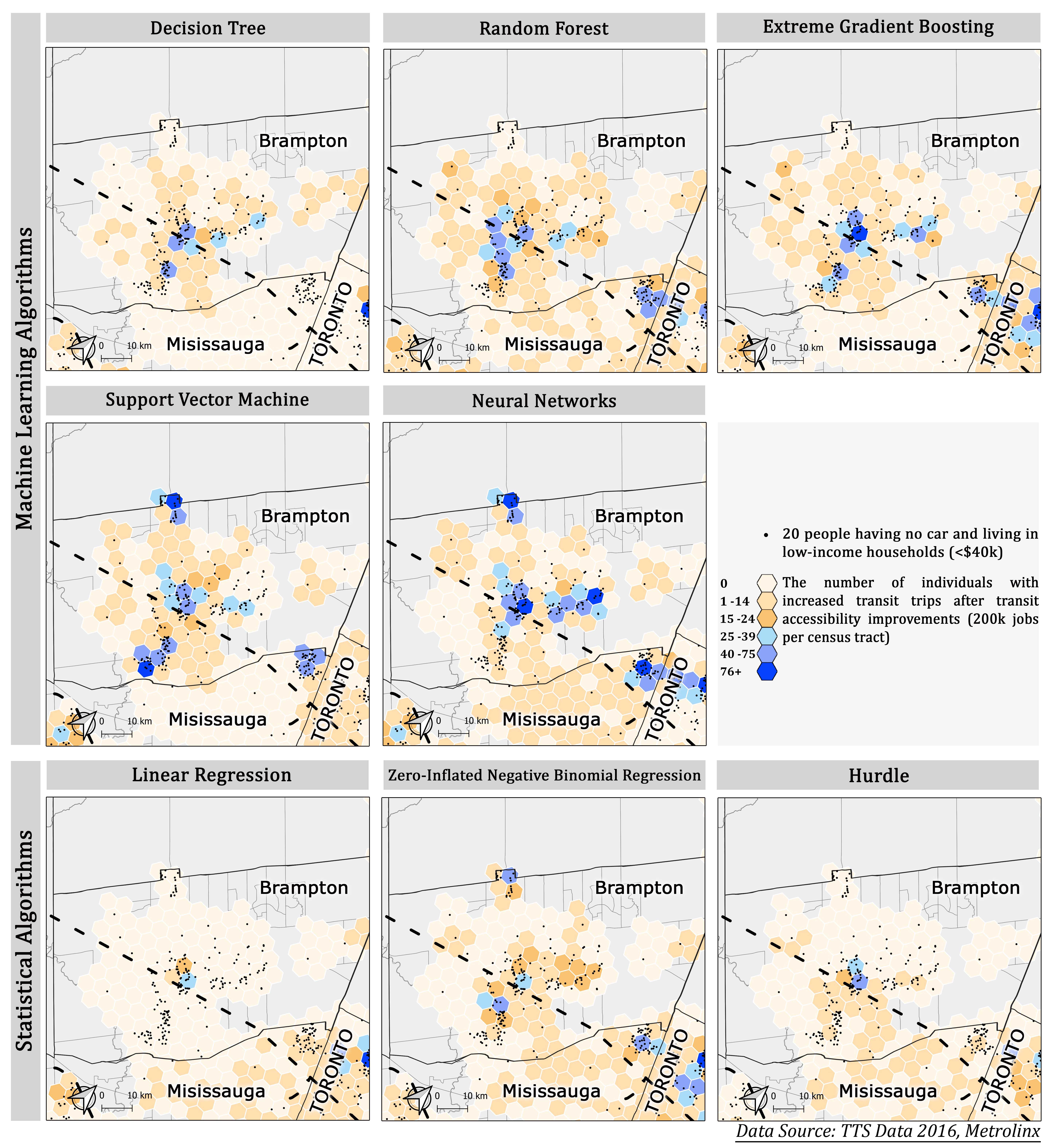}
    \caption{\textcolor{black}{The spatial prediction of newly generated transit trips in Brampton by low-income carless group after improving job accessibility (enhancement the accessibility from 0 to 200k jobs per census tract) using different algorithms}}
    \label{fig: sensitivity_bram}
\end{figure}

\textcolor{black}{Figure~\ref{fig: sensitivity_bram} shows the spatial distribution of newly generated transit trips in the city of Brampton.
The interesting point for Brampton is that although ML models generally suggest lower sensitivity of the low-income households to transit improvement compared to statistical ones (See Figures~\ref{fig: sensitivity_all} and Figure~\ref{fig: sensitivity}), in Brampton, it is the other way around. We can clearly see a higher number of new transit trips suggested by the ML models after transit enhancement in this region. Most of these new transit trips belong to downtown Brampton and its city center. The results show that ML models predict significant new trips along a corridor stretching from downtown Brampton to the east (and north and south for SVM and NN) after transit investments. On the other hand, statistical models barely suggest new transit trips by the low-income carless community. } 

\textcolor{black}{One potential reason that the ML results look so different from the statistical model results in the Brampton area is that the regional-scale dataset and models employed are dominated by observations outside of the Brampton area. It appears that the ML algorithms are better able to make more nuanced local forecasts given their non-linear structures. Future work is needed to investigate further. For example, models estimated on just the Brampton subset of the data may be compared, or the use of spatially varying models such as GWR can be explored. Be that as it may, in planning practice, much modelling occurs at the regional scale, and employing methods that can provide more locally nuanced results will always be desirable.}

\begin{figure}[!ht]
\centering
    \begin{subfigure}{.8\textwidth}
        \centering
        \includegraphics[width=1\textwidth]{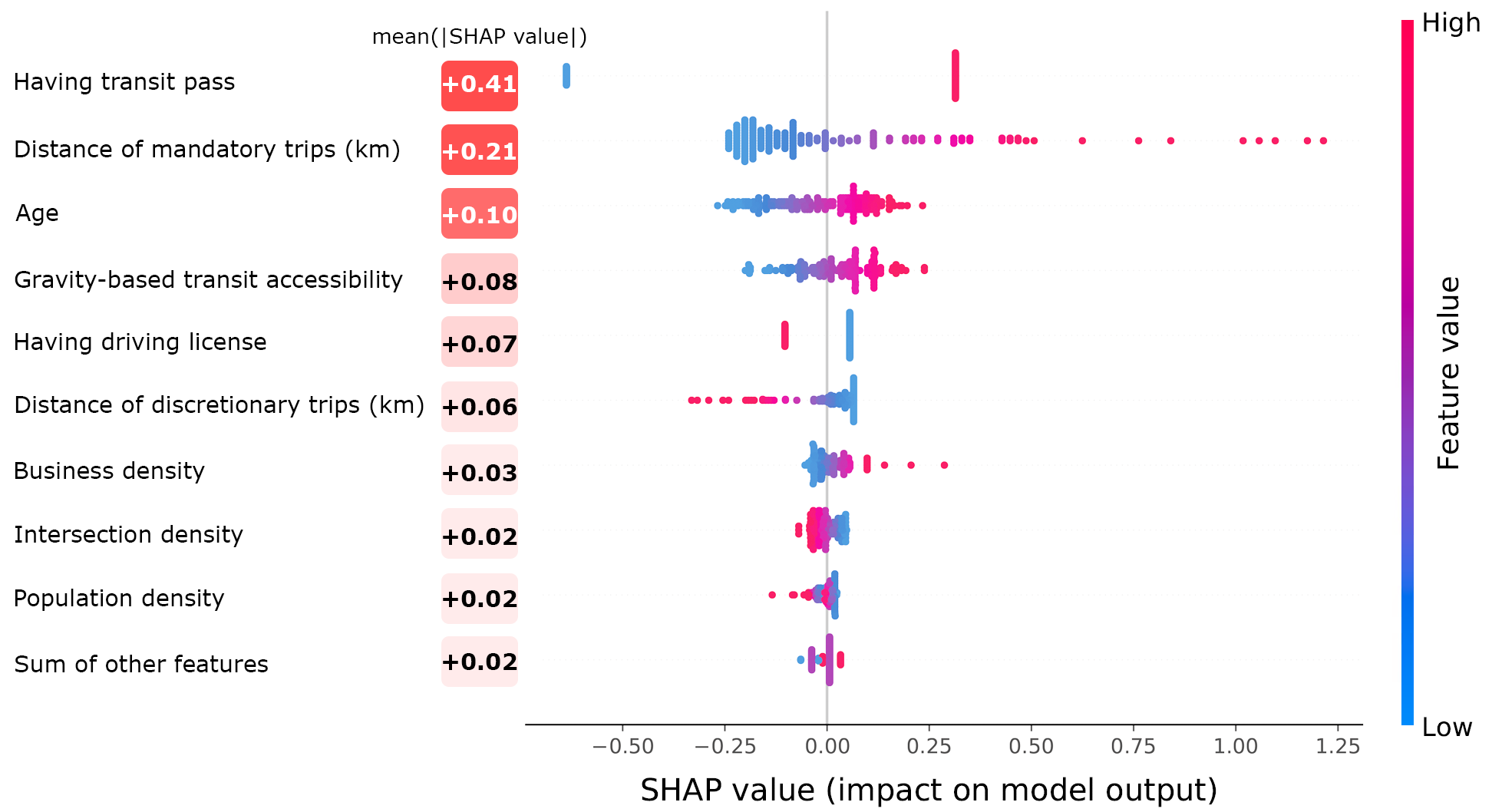}
        \caption{Linear Regrsession}
        \label{fig: LIME-LR}
    \end{subfigure}
    \\
    \begin{subfigure}{.8\textwidth}
        \centering
         \includegraphics[width=1\textwidth]{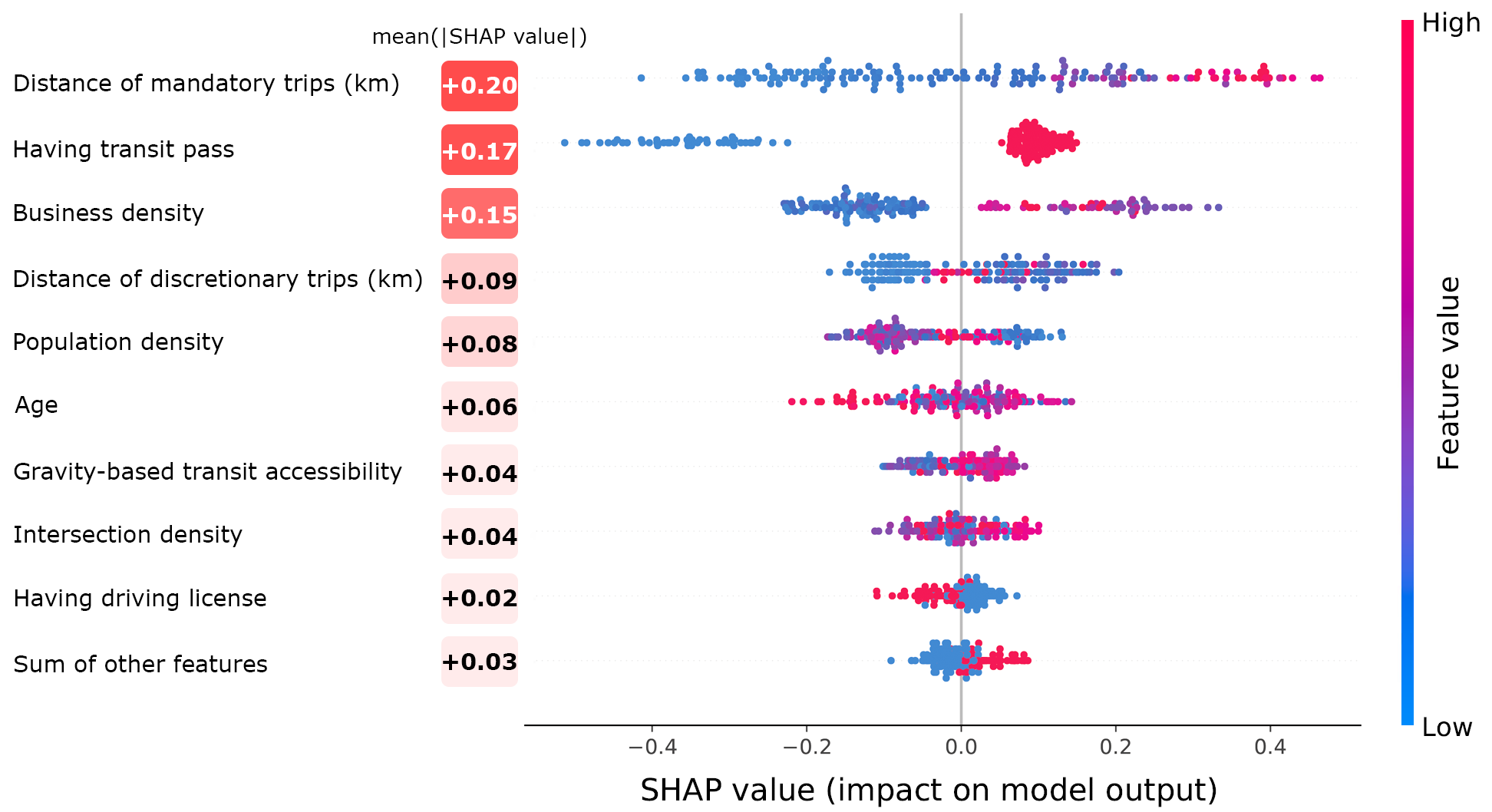}
         \caption{Random Forest}
         \label{fig: LIME-RF}
    \end{subfigure}
     \caption{\textcolor{black}{Comparing the importance of each independent variable based on their SHAP values (Low-income carless households in Brampton)}}
     \label{fig: SHAP-Comp}
\end{figure}

\textcolor{black}{We further explore the difference between ML and statistical models by considering the important factors in their prediction. Figure~\ref{fig: SHAP-Comp} juxtaposes the ranking and importance of features based on their SHAP values. We use RF and LinR algorithms to explore the change in the transit use of the low-income carless households of Brampton after transit enhancements. Then, we utilize the SHAP technique to see which independent features affect their transit use. Each model suggests different effects of independent factors in individuals' decisions. For instance, the binary variables of ``transit pass'' and ``driver license'' possession are well-separated based on the linear model, whereas RF shows a more complicated behaviour for the transit pass possession. 
Compared to the actual case, we may conclude that the transit pass does not have the same value for each person; thus, the LinR fails to capture the complexity of that attribute and its possible interaction with other features. Interestingly, in the LinR case, we observe no outlier in the predictions, i.e., the red and blue colours for two extreme points are always well-separated, no matter what the feature is. On the other hand, we can see a mixture of colours in most of the RF's attributes. In practice, we may always expect some people who can be considered outliers when it comes to decision-making. Therefore, having a pure separation does not seem an appropriate representation of the complex real world. Feature ranking also differs for each model. For instance, the population and business density seem more important to RF than LinR when predicting the response of individuals to transit improvement. Accordingly, a planner with equity concerns may decide to invest in higher density regions while utilizing RF, whereas using LinR may not arrive at the same conclusion. We recommend employing different ML and statistical algorithms in equity decisions, comparing their results, and consulting with planners and decision-makers. Therefore, there is no best model for all problems, and each model should be investigated and interpreted before being taken in the equity context.}

%%%%%%%%%%%%%%%%%%%%%%%%%%%%%%%%%%%
%%%%%%%%%%% CONCLUSION %%%%%%%%%%%%
%%%%%%%%%%%%%%%%%%%%%%%%%%%%%%%%%%%
\section{Conclusion and Limitations}
An accurate travel demand model for travel behaviour responses and then policy-making of transportation investments in low-income communities is of concern to planners. A just transit investment aiming to address environmental, economic, and social objectives requires a clear understanding of travel needs and travel responses of different populations. Hence, having a model with higher predictive accuracy is as critical as evaluating individuals' travel behaviour via an interpretable model. Policy-makers rely on a model that is easy to understand and whose variable importance is determined. Hence, to build a model accurately predicting the transit use of low-income individuals, we first presented a systematic comparison of well-established ML and statistical regressors. Then, we discussed the sensitivity and interpretability of the best-performing model.

Comparing the predictive performance of all models showed that Random Forest is the most accurate method for modelling the transit demand of low-income households. 
The value of R-squared for RF, on average, was significantly higher compared to all other statistical models. 
From the equity-based perspective, we further examined the impact of transit investment (i.e., accessibility improvement) on the potential transit trips by low-income individuals. The sensitivity of the models to the accessibility gains was significantly different. It showed a 17\% difference in the number of predicted new transit trips across the models tested. Undoubtedly, any transit plans or policies framed by each model will have different equity impacts on low-income communities. Afterward, we mapped the newly generated transit trips across all models to examine the spatial distribution of transit trips in the region. We observed a heterogeneous spatial distribution of new transit trips by vulnerable groups among traditional and ML models. For instance, ML models proposed a potential for the new transit trips in Brampton city -- i.e., favourable investment in that region. However, statistical models did not demonstrate a significant number of transit trips for the same region. \textcolor{black}{We further investigate such a difference in the ML prediction based on the city's long-term goals toward an equitable transit network. Our experiment shows that using the ML models would give rise to different policy decisions in the Brampton region compared to the statistical model. In this special case, the improved local accuracy in Brampton predicts larger increases in transit use and, thus, the benefit of increasing accessibility in Brampton. We note that, in practice, such differences should be investigated case by case.
}

We further explored the global and local interpretability of the best-performing model, i.e., the Random Forest (RF). We utilized five different model-agnostic tools to investigate the effect of each feature on the number of predicted transit trips in two levels of granularities -- e.g., group and individual interpretability. The length of the trips, having transit pass, and accessibility to public transit have the most impact on transit use. Throughout our analysis, \textcolor{black}{we found ML models more accurate and at the same time explainable via external interpretability tools}. Based on the experiment, we recommend utilizing a model that is both accurate and rational -- that is, its global and local interpretation can be supported by pre-existing knowledge or theory. 

\textcolor{black}{The results of this study have significant implications for planning, policy, and travel demand modelling. As decision-makers are increasingly looking for ways to alleviate inequalities in access to transit and improve the activity participation of households living in low-income communities, this study can help agencies examine transit investment projects and transit-related policies. This framework demonstrates the possibility of using ML methods to enhance travel demand predictions. Still, the big question is which model should be used in practice. There is a trade-off between accuracy and interpretability. In any case, we recommend the following pipeline: first, to train different ML and statistical models; second, to statistically compare the result of each algorithm; third, to select an intrinsic interpretable model if the performance difference is negligible or to choose a more complex model when the difference is significant; fourth, to explain the best model using different interpretability tools and discuss its interpretation with an expert; and fifth, to rely on the model if its result is justified by the literature and empirical interpretation, and otherwise, to use an intrinsically interpretable model.}

It is worth mentioning that there is no single remedy to model all datasets. Future studies may replicate the experiment in new regions. \textcolor{black}{Further, our research approach may also be extended by simultaneously integrating ML and spatial analysis models. Such a hybrid model may alleviate the interpretability of a single black-box algorithm.} Nevertheless, this study does not aim to propose the best model to predict travel behaviour, but rather it sheds light on the significance of model selection based on predictive performance and interpretability. On the other hand, we used the latest travel survey in the GTHA related to 2016. However, the COVID-19 pandemic has massively changed the lives of people and their travel behaviour, particularly their transit use, due to the health risks. Therefore, there is a need to investigate models that best predict longitudinal behaviour changes and compare transit use of vulnerable strata before and after accessibility changes. 

%%%%%%%%%%%%%%%%%%%%%%%%%%%%%%%%%%%
%%%%%%%% ACKNOWLEDGEMENT %%%%%%%%%%
%%%%%%%%%%%%%%%%%%%%%%%%%%%%%%%%%%%
\section*{Acknowledgements}
We are grateful to the University of Toronto Data Management Group who generously provided access to TTS data via their remote desktop servers.

%%%%%%%%%%%%%%%%%%%%%%%%%%%%%%%%%%%
%%%%%%%%% BIBLIOGRAPHY %%%%%%%%%%%%
%%%%%%%%%%%%%%%%%%%%%%%%%%%%%%%%%%%
% \bibliographystyle{elsarticle-num-names}
\bibliographystyle{elsarticle-harv}
\bibliography{reference}

%%%%%%%%%%%%%%%%%%%%%%%%%%%%%%%%%%%
%%%%%%%%%%%% APPENDIX %%%%%%%%%%%%%
%%%%%%%%%%%%%%%%%%%%%%%%%%%%%%%%%%%
\newpage
\appendix
\section{\textcolor{black}{Algorithmic representation of interpretability tools}}

\textcolor{black}{In this section, we provide the step by step implementation of each interpretability tool.}

%%%%%%%%%%%%%%%%%%
\subsection{\textcolor{black}{Feature importance algorithm}} \label{sec:feature_importance_algorithm}
\textcolor{black}{Algorithm~\ref{alg:Feature importance algorithm} describes a model-agnostic permutation-based feature importance technique introduced by~\citet{Fisher2019}. In this study, we compute the permutation-based feature importance of RF algorithm using \texttt{scikit-learn} package in \texttt{Python} platform.
}

\begin{algorithm}[!ht]
\caption{Feature importance algorithm}\label{alg:Feature importance algorithm}
\KwData{Trained model $\hat{f}$, feature matrix $X$, outcome $y$, error measure $L(y,\hat{f})$.}
Fit the model on a train data with real features and calculate the actual model performance (e.g. RMSE for a regression model);

\For{$\text{Each feature~} j \in \{1, \dots, p\}$}{
Permutate values of feature $j$ and generate a new feature matrix;

Fit the model on the modified data and estimate the permutated model performance;

Compute the difference between the actual model performance and the permutated model performance
}

Rank features according to the differences between their permutated model and the actual one

\end{algorithm}

%%%%%%%%%%%%%%%%%%
\subsection{\textcolor{black}{Partial Dependence Plot (PDP) algorithm}} \label{sec:PDP_algorithm}
\textcolor{black}{We divide the feature space $x$ into subgroups $j$ and $C$. $j$ includes the feature on which the partial dependence function $\hat{f}_j$ is applied, and $C$ corresponds to the remaining attributes in the dataset. $x_j$ and $x_C$ define the values of features in $j$ and $C$, respectively. The partial dependence function estimates the relationship between $x_j$ and the targeted variable by keeping the feature values in subgroup $C$ unchanged. Therefore, we have a function depending only on feature $j$ and the average effect of other features in $C$~\citep{Casalicchio2018}.}

\textcolor{black}{The partial dependence function $\hat{f}_j$ on $x_j$ is}
\begin{align}
    \hat{f}_j(x_j)&= \mathbb{E}_{X_C}[\hat{f}(x_j, X_C)]&& x_s\in j, x_c\in C \label{eq:PDP} \\
    \hat{f}_j(x_j)&= \frac{1}{n}\sum_{i=1}^n\hat{f}(x_j, x^{(i)}_C).
    \label{eq:ICE}
\end{align}

\textcolor{black}{Accordingly, PDP can be constructed using Algorithm~\ref{alg:PDP algorithm} as follows.}

\begin{algorithm}[!ht]
\caption{PDP algorithm}\label{alg:PDP algorithm}
\KwData{Feature matrix $X$, Set of unique values $\{x_{j1}, \dots, x_{jk}\}$ in feature $j$.}
Select feature $j$ on which the partial dependence function is applied ($C$ is the set of the remaining features).

\For{$\text{Each feature~} j \in \{1, \dots, p\}$}{
    \For{$\text{Each unique value~} x_{ji} \in \{x_{j1}, \dots, x_{jk}\}$}{
    Replace all original values $x_j$ of the selected feature $j$ with the constant value $x_{ji}$ ($i$ is the number of observations);
    
    Keep the values $x_C$ of complement features $C$ unchanged;
    
    Fit the model and compute the predicted response $\hat{f}_j$ for the modified dataset;
    
    Average the predicted value to obtain $\bar{f}_j(x_{ji})$;
    }
    
    Plot the pairs $\{x_{ji}, \bar{f}_j(x_{ji})\}$.
}
\end{algorithm}

%%%%%%%%%%%%%%%%%%

\subsection{\textcolor{black}{Individual Conditional Expectation (ICE) algorithm}} \label{sec:ICE_algorithm}

\textcolor{black}{Similar to the PDP algorithm, ICE iterates on the set of unique values for each feature; however, it plots the output per instance instead of averaging it for all observations. Accordingly, $n$ estimated response curves each of which corresponds to the value of $i$-th observation $x_j^{(i)}$, the prediction of $i$-th observation $\hat{f}_j^{(i)}(x_j^{(i)})$ while the value of the $i$-th instance for the features in $x_C^{(i)}$ is unchanged, are plotted. Therefore, ICE plots include curves $\hat{f}_j^{(i)}$ for each observation in $\{(x_j^{(i)},x_C^{(i)})\}_{i=1}^n$~\citep{Molnar2020,Casalicchio2018}. In other words, the ICE plot is a disaggregated form of PDP.}

\subsection{\textcolor{black}{Local interpretable model-agnostic explanations (LIME) algorithm}} \label{sec:LIME interpretability}

\textcolor{black}{In order to demonstrate LIME interpretation in our study, we follow these steps:}

\begin{enumerate}
    \item \textcolor{black}{We select the person for whom we want an explanation for the black-box model (e.g., Neural Networks).}
    \item \textcolor{black}{We perturb the dataset to get the predictions for the black-box model using these new points.}
    \item \textcolor{black}{We weight the new perturbed samples based on how close they are to the person.}
    \item \textcolor{black}{We train a weighted, interpretable model (Linear Regression in our case) on the new dataset.}
    \item \textcolor{black}{We explain the prediction by analysing the local model.}
\end{enumerate}

\textcolor{black}{As we have a regression experiment, we employ the linear regression model as the interpretable model in LIME.}

%%%%%%%%%%%%%%%%%%%%%%%%%%%%%%%%%%%%%
%%%%%%%%%%%%%%%%%%%%%%%%%%%%%%%%%%%%
\section{Detailed comparison of models' performances} \label{sec:detailed_comp}

Table~\ref{tab:detailed_comp_regress} shows a detailed statistical comparison of regressors' performances. It lists the best-performing regressor and statistically tied models. The average ranks of the models based on each metric are written within parentheses. According to the Bergmann-Hommel post hoc test, ML models, including tree-based algorithms and NN, are the best models for estimating the number of transit trips for vulnerable groups. Moreover, traditional models, e.g., ZINB and Hurdle, have the lowest predictive power. Accordingly, we recommend utilizing ML algorithms to model a regression travel-mode problem. 

\begin{table}[!ht]
\centering
\caption{Comparing the performance of the best regressors and the baseline regressors using the Friedman Aligned Ranks test and its post hoc analysis.}\label{tab:detailed_comp_regress}
\resizebox{0.85\linewidth}{!}{
    \begin{tabular}{lllc}
    \toprule
        \textbf{Metric} &
          \textbf{\begin{tabular}[c]{@{}l@{}}Friedman Aligned Ranks\\ test ($p$-value)\end{tabular}} &
          \textbf{\begin{tabular}[c]{@{}l@{}}The best model\\ and possible ties\end{tabular}} &
          \textbf{\begin{tabular}[c]{@{}l@{}}Traditional\\ models' mean rank\end{tabular}} \\
          \midrule
            \textbf{R\_Squared} &
          1.18e-11* &
          \begin{tabular}[c]{@{}l@{}}b: RF  (1)\\ t: NN  (2.2) \& XGB (3.2)\end{tabular} &
          \begin{tabular}[c]{@{}l@{}}LinR (8) \& ZINB (6)\\ \& Hurdle (7)\end{tabular} \\ [0.4cm]
        \textbf{RMSE} &
          1.08e-01* &
          \begin{tabular}[c]{@{}l@{}}b: RF  (1)\\ t: NN  (2.2) \& XGB (3.2)\end{tabular} &
          \begin{tabular}[c]{@{}l@{}}LinR (8) \& ZINB (6)\\ \& Hurdle (7)\end{tabular} \\ [0.4cm]
        \textbf{MedAE} &
          8.76e-12* &
          \begin{tabular}[c]{@{}l@{}}b: DT  (1)\\ t: SVM (2.1) \& NN (3.1)\end{tabular} &
          \begin{tabular}[c]{@{}l@{}}LinR (8) \& ZINB (4.1)\\ \& Hurdle (7)\end{tabular} \\ [0.4cm]
        \textbf{RRSE} &
          1.06e-11* &
          \begin{tabular}[c]{@{}l@{}}b: RF  (1)\\ t: NN  (2.2) \& XGB (3.2)\end{tabular} &
          \begin{tabular}[c]{@{}l@{}}LinR (8) \& ZINB (6)\\ \& Hurdle (7)\end{tabular} \\
        \bottomrule
        \multicolumn{4}{l}{\begin{tabular}[c]{@{}l@{}}\qquad b: the best model; \quad t: possible ties with insignificant difference;\\  \qquad  *: statistically significant based on $\alpha = 0.05$\end{tabular}} \\
    \end{tabular}
}
\end{table}

\end{document}